\documentclass[runningheads]{llncs}
\usepackage[T1]{fontenc}
\usepackage{graphicx}
\usepackage{booktabs}
\usepackage[misc]{ifsym}
\newcommand{\corr}{(\Letter)}

\usepackage{cite}
\usepackage[hidelinks]{hyperref}
\usepackage[utf8]{inputenc}
\usepackage{amsmath,amsfonts,amssymb}
\usepackage[ruled,vlined]{algorithm2e}
\graphicspath{{./}{figures_ijcai/}}
\SetKwInput{KwIn}{Input}
\SetKwInput{KwResult}{Output}
\SetAlgoLined
\DontPrintSemicolon
\newtheorem{assumption}{Condition}

\newcommand{\cN}{\mathcal{N}}
\newcommand{\cB}{\mathcal{B}}
\newcommand{\cQ}{\mathcal{Q}}

\newcommand{\cD}{\mathcal{D}}

\newcommand{\err}{\mathrm{Err}}

\newif\ifwithappendix
\withappendixfalse

\begin{document}

\title{COSMOS: Model-Agnostic Personalized Federated Learning with Clustered Server Models and Pseudo-Label-Only Communication\thanks{This version was published at ECML PKDD 2026.}}
\titlerunning{COSMOS: Model-Agnostic Personalized Federated Learning}

\author{
Ben Rachmut\inst{1}
\and Luise Ge\inst{1}\corr
\and Ning Zhang\inst{1}
\and William Yeoh\inst{1}
\and Yevgeniy Vorobeychik\inst{1}
}
\authorrunning{B. Rachmut et al.}
\tocauthor{Ben Rachmut, Luise Ge, Ning Zhang, William Yeoh, Yevgeniy Vorobeychik}
\toctitle{COSMOS: Model-Agnostic Personalized Federated Learning with Clustered Server Models and Pseudo-Label-Only Communication}

\institute{Washington University in St. Louis \email{g.luise@wustl.edu}}

\maketitle

\begin{abstract}
Federated learning (FL) in heterogeneous environments remains challenging because client models often differ in both architecture and data distribution. While recent approaches attempt to address this challenge through client clustering and knowledge distillation, simultaneously handling architectural and statistical heterogeneity remains difficult. We introduce COSMOS, a model-agnostic framework that enables server-side personalization using only pseudo-label communication. Clients train local models and predict on the public data; the server clusters clients by prediction similarity, trains a cluster-specific model for each group using its own compute, and distills the resulting models back to clients. We provide the first theoretical analysis showing that distillation from the learned cluster models can yield exponential personalization risk contraction, going beyond the convergence-to-stationarity guarantees typically provided in model-agnostic FL. Experiments across benchmarks demonstrate that COSMOS consistently outperforms all model-agnostic FL baselines while remaining competitive with state-of-the-art personalized FL methods. More broadly, our results highlight personalized server-side learning with pseudo-labels as a promising paradigm for scalable and model-agnostic federated learning in highly heterogeneous environments.


\end{abstract}

\section{Introduction}

Federated learning (FL) is a distributed training paradigm where clients collaboratively train one or more server-side models without disclosing local data~\cite{mcmahan2017communication,li2020federated,karimireddy2020scaffold}.
Motivated by the heterogeneity of client data distributions, a host of personalized FL (PFL) methods has been developed to tailor models to individual clients~\cite{li2019fedmd,cho2023communication,abourayya2025little,tan2022towards}.
However, most existing FL schemes still presume some structural knowledge or compatibility of client model architectures.
This imposes a significant practical barrier, as clients may often wish to use whatever model architecture best fits their needs, or make use of proprietary architectures that they would not wish to disclose.
Consequently, an important practical need in FL is to be \emph{model agnostic}, allowing clients using whatever models they choose to simply ``plug in'' to the FL scheme. The implication of model-agnostic FL is that it disallows any communication about model parameters or gradients.

Despite the clear practical need, the problem of model-agnostic FL remains underexplored, particularly when clients simultaneously differ in both data distributions and model architectures. Existing model-agnostic approaches therefore rely on output-level communication, typically utilizing a shared unlabeled dataset~\cite{li2019fedmd,cho2023communication,abourayya2025little}. Such datasets provide a common reference set on which heterogeneous models can exchange predictive signals without revealing parameters or private data. In many practical deployments, such datasets are readily available through public corpora (e.g., web-scraped images or text), synthetic generation, or institutionally shared benchmark pools. As a result, prediction-based communication has emerged as one of the most practical mechanisms for enabling collaboration across heterogeneous models in FL. Moreover, by using pseudo-labels instead of parameters or gradients, communication efficiency can be improved significantly\ifwithappendix\ (see our comparison in Table~\ref{table:communication_comparison})\fi, which has been widely recognized as a critical concern in FL as the wireless and other end-user connections are typically slower, more expensive and less reliable~\cite{shahid2021communication}.

To the best of our knowledge, COMET~\cite{cho2023communication} is the closest prior work that targets the crossover. However, COMET has three major limitations. First, the server acts solely as a passive coordinator, neglecting the significant computational resources potentially available at the server side that could be leveraged to facilitate personalization. Second, its reliance on heuristic K-means clustering necessitates prior knowledge of the cluster count K and lacks a formal mechanism for personalization when the underlying client diversity is high. Consequently, COMET's theoretical framework is restricted to standard convergence-to-stationarity for non-convex objectives, providing no formal guarantees regarding personalization performance or risk reduction.

We address all these limitations with COSMOS (\textbf{C}lustered \textbf{O}utput-based \textbf{S}erver \textbf{Mo}del\textbf{s}). While a high-capacity server has the potential to assist clients, it is highly non-trivial whether the server can effectively learn from the noisy, heterogeneous pseudo-labels provided by the clients in the first place. COSMOS explicitly overcomes this bottleneck through careful algorithmic design. Specifically, COSMOS enables clients to train arbitrary local models on their private data and use them to generate pseudo-labels on a shared unlabeled dataset. The server then performs distance-controlled clustering to group clients with similar data distributions and trains a dedicated teacher model for each cluster. Notably, COSMOS does not require the dataset to match client distributions exactly, only that it provides broad coverage of the input space. As we show in the experiments (Section~\ref{S:exp}), COSMOS remains effective even when the public dataset constitutes only a small fraction of the overall training data. At the same time, the quality of this public pool remains a genuine limiting factor: when coverage is poor or the public/private distribution shift is severe, both clustering quality and pseudo-label reliability can degrade, so public-data construction is an important practical design choice rather than a free assumption.

Another critical gap in all existing model-agnostic PFL literature is the absence of risk contraction guarantees. To address this, we provide the first end-to-end analysis of personalization risk contraction. To achieve this level of rigor, we leverage standard tools from semi-supervised learning (SSL) theory \cite{lang2024theoretical,Wei2020TheoreticalAO}, including expansion-based connectivity and bounded pseudo-label error. They allow us to derive general sufficient conditions for exponential risk contraction without restricting the model class. Importantly, these assumptions are used only for analysis and do not impose constraints on the practical implementation of COSMOS.

Our main contributions are:
\begin{enumerate}
    \item \textbf{Algorithmic Framework.} We present COSMOS, the first model-agnostic PFL framework where the server actively trains cluster-specific models using clients' pseudo-labels.
    
    \item \textbf{Theoretical Guarantee.} We establish an end-to-end exponential contraction of personalization risk bounds for COSMOS under sufficient conditions, 
    providing the first general risk contraction guarantee in model-agnostic PFL. 
           \item \textbf{Empirical Evaluation.} We demonstrate that COSMOS not only consistently outperforms existing model-agnostic FL methods but also maintains competitive performance in homogeneous settings, while reducing communication from parameter sharing by 1-2 orders of magnitude.\footnote{Our code is publicly available at \url{https://github.com/YODA-Lab/COSMOS}}

\end{enumerate}

\section{Related Work}
\label{S:relwork}

\noindent \textbf{Classical Federated Learning.} Federated learning was introduced through FedAvg~\cite{mcmahan2017communication}, which trains a single global model by aggregating client weight updates. While simple and communication-efficient, FedAvg suffers with non-IID data, motivating methods such as FedProx~\cite{li2020federated} and SCAFFOLD~\cite{karimireddy2020scaffold} that stabilize optimization via proximal or control-variance corrections. Nonetheless these methods converge to a single global model and do not offer personalization.

\smallskip
\noindent \textbf{Model-Agnostic, Model-Heterogeneous, and Knowledge-Distillation--Based Federated Learning.}
Since classical federated learning communicates parameters or their updates, it requires every model on the clients and the server to share the same architecture. 
To provide greater flexibility, a number of model-heterogeneous approaches, such as communicating instance-level representations as in FedHeNN~\cite{makhija2022architecture}, or abstract class prototypes as in FedProto~\cite{tan2022fedproto}, have been proposed to relax the architectural homogeneity assumption, as have many personalized FL methods (see below). However, while model-agnostic approaches are necessarily model-heterogeneous, most model-heterogeneous methods are \emph{not} model-agnostic, since they still impose some architectural constraints. Additionally, model-agnostic FL is necessarily knowledge-distillation--based FL (KD-FL)~\cite{mora2024knowledge}.
Nevertheless, many KD-FL methods still rely on parameter aggregation at certain stages~\cite{lin2020ensemble,sattler2020communication,zhu2021data,afonin2021towards}.
To our knowledge, only FedMD~\cite{li2019fedmd} and COMET~\cite{cho2023communication} explore purely model-agnostic FL using soft labels, while FedCT~\cite{abourayya2025little} relies on hard labels. Recently, the  communication efficiency of pseudo-labels has also been leveraged in federated multi-view clustering (e.g., CeFMC~\cite{Liu2025CommunicationEfficientFM}) although its objective is orthogonal to ours.

\smallskip
\noindent \textbf{Personalized Federated Learning (PFL).} Almost all practical federated learning settings exhibit statistical heterogeneity, where different clients' local distributions can vary substantially. Personalized federated learning (PFL) addresses this by learning client-adapted models~\cite{tan2022towards}. Existing PFL approaches can be broadly grouped by whether they maintain a single shared server model or a small number of server-side models. In the first group, a single global model is adapted to each client via meta-learning (Per-FedAvg \cite{fallah2020personalized}), regularization (pFedMe \cite{t2020personalized}, Ditto \cite{li2021ditto}), adaptive mixing of local and global models (APFL \cite{deng2020adaptive}), representation refinement (FedBABU \cite{oh2021fedbabu}), or hypernetwork-based parameter generation (pFedHN \cite{shamsian2021personalized}, FedSelect \cite{tamirisa2024fedselect}). In the second group, clustered PFL methods explicitly maintain multiple server-side models and assign clients to them, as in IFCA \cite{ghosh2020efficient}, FedGroup \cite{duan2020fedgroup}, AutoCFL \cite{gong2022adaptive}, and pFedCK \cite{zhang2024personalized}.
To the best of our knowledge, only COMET is also a model-agnostic PFL approach~\cite{cho2023communication}. However, it requires one to specify the number of clusters $K$ in advance, and applies standard $K$-means clustering, which is heuristic and not easily amenable to theoretical personalization guarantees. 



\smallskip
\noindent \textbf{Theory for Model-Agnostic FL.}
While theoretical convergence results abound for conventional FL schemes, model-agnostic settings as well as personalization make such results significantly more challenging.
The earliest model-agnostic FL approach, FedMD~\cite{li2019fedmd} does not provide any theoretical guarantees. 
A recent FedCT method~\cite{abourayya2025little} requires an oversimplified assumption directly that the training algorithms \emph{always} yield monotone increasing accuracy to achieve convergence.
The theoretical analysis for COMET~\cite{cho2023communication}, on the other hand, requires linear models and a Gaussian data distribution to obtain generalization results.
Thus, there are no general sufficient conditions on risk bound contraction for model-agnostic personalized FL.
We bridge this gap by adopting the analysis tools from 
the semi-supervised learning literature~\cite{Wei2020TheoreticalAO}.

\section{Model}
\label{sec:formulations}

We consider a federated learning scenario for $M$-class classification with $N$ clients and a server. 
Each client $i\in[N]$ has a private labeled dataset $D_i=\{(x_{ij},y(x_{ij}))\}_j$ with $x_{ij}$ drawn i.i.d.\ from its local distribution $\mathcal{D}_i$ over the input space $\mathcal{X}$, and $y(x_{ij})$ the true label of $x_{ij}$.
We assume that each $\mathcal{D}_i$ admits a density function $p_i(x)$.
A client $i$ trains a model $f_i:\mathcal{X} \rightarrow [0,1]^M$ with $\|f_i(x)\|_1 = 1$ from a hypothesis class $\mathcal{H}_i$ that can be distinct for each $i$, representing, for example, distinct neural network architectures for different clients.
A server, in turn, has a hypothesis class $\mathcal{H}_S$ and can train a \emph{collection} of models $H=\{h_1,\ldots,h_K\} \subset \mathcal{H}_S$, where $h_k:\mathcal{X} \rightarrow [0,1]^M$ and $\|h_k(x)\|_1 = 1$ for each $k$.
In our setting, the value of $K$ is obtained
\emph{endogenously as part of the training procedure}.
Furthermore, let $\pi:[N]\rightarrow[K]$ be a mapping (also obtained during training) which assigns each client $i$ to a corresponding server model $h_{\pi(i)}$.
Finally, following prior work on PFL, we assume availability of an \emph{unlabeled public dataset} $U=\{x_j\}$ of size $n=|U|$, with each $x_j$ drawn i.i.d.~from a global distribution $\mathcal{Q}$ over $\mathcal{X}$.
We assume that $\mathcal{Q}$ admits a density function $q(x)$.
It is not difficult to obtain unlabeled datasets of this kind, for example, scraping (open-license) images or text from the internet, or generating it synthetically.



To formalize the learning objective, we introduce some additional notation.
Let $g$ be a classification model with outputs a distribution over $M$ classes (i.e., pseudo-labels).
We use $A\circ g(x) = \arg\max_{m\in[M]} [g(x)]_m$ to denote the predicted class (i.e., the class with the highest probability under $g$).
Further, let $R_{\mathcal{D}}(g) = \mathbb{E}_{x\sim\mathcal{D}}
[
\ell_{0-1}\{A\circ g(x), y(x)\}
]
$
    denote the risk (probability of a mistake) of $g$ under distribution $\mathcal{D}$, where $\ell_{0-1}$ is the $0-1$ loss, and $\err_{x\in D}(g)= \frac{1}{|D|} \sum_x
\ell_{0-1}\{A\circ g(x), y(x)\}
$ denote the hard label prediction error over a dataset $D$.


\smallskip
\noindent\textbf{Learning Objective:} Our goal is to train a collection $\{h_k\}$ of $K$ personalized server models, along with a client-to-model mapping $\pi$, which minimizes the \emph{personalization risk}, defined as the average risk over the $N$ clients:
\(
R^\dagger(h_1,\dots,h_K)
=\frac{1}{N}
\sum_{i=1}^N R_{\mathcal{D}_i}(h_{\pi(i)}).
\)
\ifwithappendix
A full notation summary appears in Table~\ref{table:notation} in Appendix~\ref{app:notation}.
\fi



\section{Algorithmic Approach}
\label{sec:framework}

\begin{figure}[t]
\centering

\includegraphics[width=0.7\columnwidth]{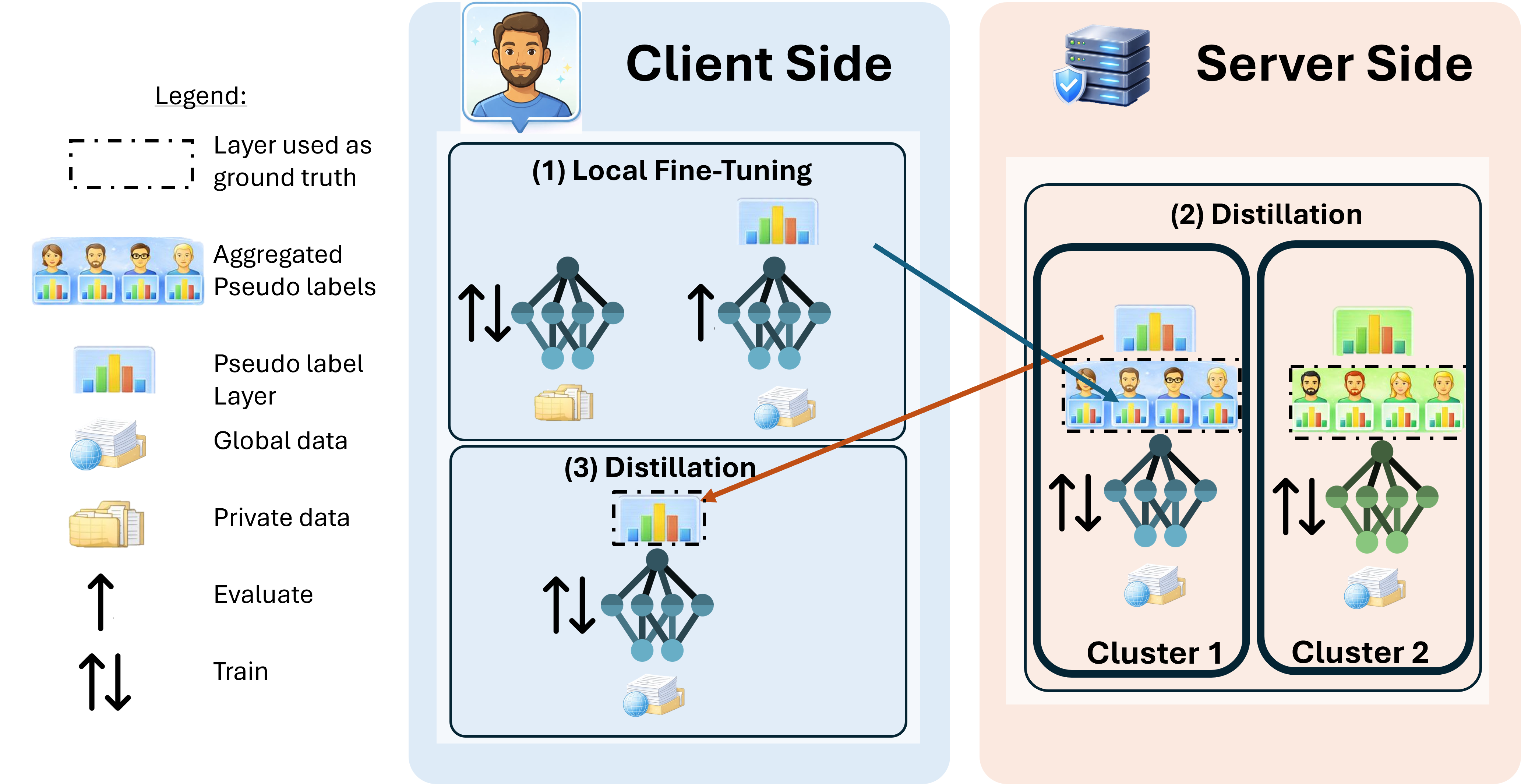}

\caption{Overview of the four steps in the COSMOS workflow.}
\label{fig:cosmos-workflow}

\end{figure}

At the high level, the proposed COSMOS framework involves iteratively fine-tuning the client and server models using pseudo-labels collected from one another. As shown in Figure~\ref{fig:cosmos-workflow}, clients generate pseudo-labels for global data using locally trained models, which the server clusters and aggregates to train cluster-specific models. The resulting pseudo-labels are iteratively returned to clients for local model refinement. We enable personalization by creating and training a small collection $H$ of $K$ server models, with each server model $h_k$ specialized to a subset of similar clients obtained through clustering.
More precisely, COSMOS proceeds in two phases: 1) the \emph{pre-training and clustering (PTC)} phase and 2) the \emph{iterative federated fine-tuning (IFFT)} phase.

\subsection{The Pre-Training and Clustering Phase}
\label{S:phase1}

The pre-training and clustering phase involves four steps: 1) local pre-training, 2) clustering,  3) server-side distillation, and 4) client-side distillation.

\smallskip
\noindent\textbf{Step 1: Local Pre-Training.}
We pre-train each client $i$'s model $f_i$ locally on $D_i$ for $E_0$ epochs using conventional (e.g., cross-entropy) loss.
We refer to the resulting client models as $f_i^{(1,1)}$, where the first superscript refers to the time step ($t=1$) and the second means that the client models are trained (or fine-tuned, in the IFFT phase) on \emph{local} data (as opposed to global data $U$ with server-provided pseudo-labels, as in Step 4 below).
Each client $i$ then generates pseudo-labels over the global set $U$, $f_i^{(1,1)}(U)$, and sends these to the server.

\smallskip
\noindent\textbf{Step 2: Client Clustering.}
Upon receiving pseudo-labels $f_i^{(1,1)}(U)$ from all clients $i$, the server proceeds to cluster the clients into $K$ clusters $\{\mathcal{G}_k\}_{k=1}^K$ based on pseudo-label similarity.
Our goal is to find the minimum number of clusters $K$ with respect to a given distance function.
However, this problem is NP-hard and
the best known approximation
has computational complexity exponential in the data dimension~\cite{ge2025learning}.
On the other hand, while numerous heuristic clustering approaches have been proposed, they cannot be easily used to provide \emph{risk convergence} guarantees.

To address this challenge, we propose the following greedy approach, which \emph{is} amenable to theoretical analysis (see Section~\ref{S:clustering}).
We define the distance between two clients $i,j$ as their pseudolabels' $\ell_1$ distance, i.e. $d^{(t)}(i,j)= |f_i^{(t,1)}(U)-f_j^{(t,1)}(U)|_1$.
Define the set $N_i(B_0,C)$ of \emph{$B_0$-close neighbors} of a client $i$ among clients in $C$ as $N_i(B_0,C) = \{j | j\in C \setminus \{i\}, d^{(1)}(i,j)\le B_0\}$.
At the high level, our algorithm greedily selects a client with the most $B_0$-close neighbors among those not previously selected, and defines a new cluster associated with this client and its neighbors, proceeding until all clients belong to some cluster.
This is made more precise in Algorithm~\ref{A:clustering}.
The resulting clustering scheme induces a cluster assignment function $\pi$ which maps each client $i$ to a cluster $k$. Notably, \emph{we need not know the number of clusters upfront}, as this is endogenous to our clustering approach. While $B_0$ can be treated as another hyperparameter for clustering, this hyperparameter is theoretically grounded: it is directly connected to the personalization guarantees of our framework. In practice, this means COSMOS is not hyperparameter-free: although $K$ is not supplied directly, the threshold $B_0$ must still be tuned, for example by validation performance or a communication/compute budget.

\begin{algorithm}[t!]
\KwIn{Clients $1,\dots,N$ with associated hard label vectors $\{y_i\}$; clustering parameter $B_0$.}
\KwResult{The number of clusters $K$ and the associated collection of client clusters $G=\{\mathcal{G}_k\}_{k=1}^K$.}

\textbf{Initialization:} $C = [1,\dots,N]$, $G = \{\emptyset\}$, $K=1$\;

\While{$C \ne \emptyset$}{
    $i^* \in \arg\max_{i \in C}$ $|N_i(B_0,C)|$\;
    $\mathcal{G}_K = \{i^* \cup N_{i^*}(B_0,C)\}$\;
    $G = G \cup \mathcal{G}_K$; $C = C \setminus \mathcal{G}_K$; $K = K+1$\;
}
\caption{Greedy Clustering.}
\label{A:clustering}
\end{algorithm}

\smallskip
\noindent\textbf{Step 3: Server-Side Distillation.}
Our next step is to train the server-side model in each cluster $k$ based on the pseudolabels received from all clients.
Specifically, fix a cluster $\mathcal{G}_k$ and a datapoint $x \in U$.
The server update for each cluster model $h_k^{(1)}$ (where the superscript references iteration) and $x \in U$ then uses the aggregate pseudo-labels over all clients in this cluster 
which we define as $\bar{f}^{(1)}_k(x)=\frac{1}{|\mathcal{G}_k|}\sum_{i \in \mathcal{G}_k} f_{i}^{(1,1)}(x)$ for clients $i \in \mathcal{G}_k$, i.e., the average of the pseudo-labels of the clients in the $k$-th cluster.
Finally, the server performs gradient-based training for each cluster-specific model $k$ to minimize the following objective:
    \begin{align}
    \label{E:server_update}
    \!\!\!\!\!\!\hat J(h_k^{(1)};\ell)
    =
    \sum_{x\in U}
    \Big[
        \ell(h_k(x),\bar{f}_k^{(1)}(x))
        + \lambda\,r_\mathcal{B}(h_k^{(1)};\ell)(x)
    \Big],
    \end{align}
where $\ell(\cdot,\cdot)$ is a loss function and $r_{\mathcal{B}}(\cdot, \ell)$ an input-specific regularization term.
The role of the regularization term in our setting is to train the models which are locally smooth (related to the consistency and robustness properties central to our theoretical analysis in Section~\ref{S:theory}).
Specifically, let $\mathcal{T}$ denote a set of permissible data augmentations (e.g., image translations and rotations) and fix $d$ to be a small radius capturing minor perturbations around augmented data points (in particular, we assume that $d$ is significantly smaller than the typical norm of $x$). 
For an input $x \in \mathcal{X}$, define its \textit{transformation ball} as:
$\mathcal{B}(x) = \{x'\in \mathcal{X} : \exists T \in \mathcal{T} \text{ s.t. } \|x' - T(x)\| \leq d\}$. 
Next, let the neighborhood $\mathcal \cN(x)$ of a point $x$ be the set of inputs whose transformation balls intersect with $\mathcal B(x)$:
$\mathcal N(x) = \{ x' : \mathcal B(x) \cap \mathcal B(x') \neq \emptyset \}$. 
Let $S \subset \mathcal{N}(x)$ be a finite sample of the neighborhood of $x$ (e.g., obtained by rejection sampling).
We now define a regularization term for any model $g$ as
$r_\mathcal{B}(g;\ell)(x) := \sum_{x'\in S}\bigl\{ \ell ( g(x'),g(x))\bigr\}$. We use the same loss function for the supervised term and regularized term, but allowing different loss functions is also admissible. 
We motivate this form of regularization more precisely in Section~\ref{S:theory}.
This server-side stage is also the main additional systems cost of COSMOS relative to passive-server methods such as COMET: the total server work scales with the number of clusters, the size of the public pool, and the chosen server architecture, although the cluster models can be trained in parallel when compute is available.

After gradient-based training for each server-side model $h_k^{(1)}$ using the objective~\eqref{E:server_update}, we send the resulting pseudo-labels $h_k^{(1)}(U)$ to clients in cluster $\mathcal{G}_k$.

\smallskip
\noindent\textbf{Step 4: Client-Side Distillation.}
Finally, each client $i$ fine-tunes its local model $f_i^{(1,2)}$, where the second superscript refers to the second round of local fine-tuning based on the pseudo-labels $h_{\pi(i)}^{(1)}(U)$.
We use the following objective:
\begin{equation}
    \label{E:client_update}
\hat{F}(f_i^{(1,2)};\ell)
=
\sum_{x \in U} w_i(x)
\Big[
\ell(f_i^{(1,2)}(x),h_{\pi(i)}^{(1)}(x)) \\
+ \lambda\,r_{\mathcal{B}}(f_i^{(1,2)};\ell)(x)
\Big],
\end{equation}
where $w_i(x)$ are the client-specific weights of data in $U$ that allow for importance sampling techniques (see Section~\ref{S:theory}). In practice, we find that setting $w_i(x)=1$ for all $i$ and $x$ is effective in practice, and avoids the need to make assumptions about the local and global distribution differences.

\subsection{The Iterative Federated Fine-Tuning Phase}

After the PTC phase, we enter an iterative FL phase in which we alternate 1) local fine-tuning, 2) server-side distillation, and 3) client-side distillation over a fixed number of iterations.
We now describe each step for a fixed iteration $t$.

\smallskip
\noindent\textbf{Step 1: Local Fine-Tuning.}
The PTC phase effectively serves as iteration $t=1$.
In any IFFT iteration $t\ge 2$, the local fine-tuning step for client $i$ starts from the model
$f_i^{(t-1,2)}$ (obtained by fine-tuning on the public data $U$ annotated with the server's pseudo-labels, as described in Step~3 below),
and performs supervised training on its labeled local dataset $D_i=\{(x,y)\}$ to obtain $f_i^{(t,1)}$.
Concretely, as before, client $i$ approximately minimizes the empirical risk with a calibrated surrogate loss $\ell$ such as cross-entropy loss
\(
\hat R_{\ell,D_i}(f)\;=\;\frac{1}{|D_i|}\sum_{(x,y)\in D_i}\ell\!\big(f(x),y\big)).
\)
For the contraction analysis, we explicitly assume this supervised update is \emph{classification-safe}: with high probability over the local sample and optimization noise,
it does not increase the true $0$--$1$ risk on $\cD_i$ beyond the generalization slack.
After fine-tuning, the pseudo-labels $f_i^{(t,1)}(U)$ are then sent to the server to fine-tune the model $h_{\pi(i)}^t$.

\smallskip
\noindent\textbf{Step 2: Client-Side Distillation.}
In iteration $t$, each server model $k \in [1,\ldots,K]$ is fine-tuned on the pseudo-labels $f_i^{(t,1)}(U)$ received from clients $i \in \mathcal{G}_k$ in a way that closely mirrors Step 3 of the PTC phase.
That is, we define $\bar{f}^{(t)}_k(x)=\frac{1}{|\mathcal{G}_k|}\sum_{i \in \mathcal{G}_k} f_{i}^{(t,1)}(x)$ for $i \in \mathcal{G}_k$.
The server model $h_k^{(t)}$ is then fine-tuned using the objective~\eqref{E:server_update}, with $h_k^{(1)}$ replaced by $h_k^{(t)}$ and $\bar{f}^{(1)}_k(x)$ replaced by $\bar{f}^{(t)}_k(x)$.
The server then sends pseudo-labels $h_k^{(t)}(U)$ to clients $i \in \mathcal{G}_k$ for all $k$.

\smallskip
\noindent\textbf{Step 3: Client-Side Distillation.}
Finally, we fine-tune each client $i$ from the pseudo-labels $h_{\pi(i)}^{(t)}(U)$ using the objective from Equation~\ref{E:client_update} (as in Step 4 of PTC), in which we replace $f_i^{(1,2)}$ with $f_i^{(t,2)}$ and $h_{\pi(i)}^{(1)}$ with $h_{\pi(i)}^{(t)}$.

\section{Personalization Guarantees}
\label{S:theory}

In this section, we analyze the population risk of COSMOS with respect to the client distributions \(\{ \mathcal D_i \}_{i=1}^N\).
We provide such guarantees both for the individual client models, as well as for the overall \emph{personalization risk} on the server.
This analysis entails several technical challenges.
First, as clustering is NP-hard, heuristic approaches are typically used, and it is therefore not evident how to achieve convergence guarantees for personalization risk for algorithms that leverage clustering (as COSMOS does).
Second, it is clear that arbitrary distributions $\mathcal{D}_i$ (of the client data) and $\mathcal{Q}$ (of global data) cannot achieve convergence.
For example, if $\mathcal{Q}$ puts all mass on a single datapoint, the clustering step will fail.
Prior approaches deal with this issue by making strong distributional assumptions, such as assuming a Gaussian data distribution~\cite{cho2023communication}.
We aim to obtain \emph{general} sufficient conditions on the data distributions.
Third, we wish to make minimal assumptions on the nature of client and server models, unlike prior approaches that require linearity for generalization bounds~\cite{cho2023communication,shamsian2021personalized}.

\smallskip
\noindent \textbf{Notation:}
Before diving into the technical details, we define some useful notation.
For a client $i$, let $\mathcal{S}_i = \mathrm{supp}(\cD_i) \subseteq \mathcal{X}$ denote the support of $i$'s local data distribution and $\mathcal{S}_i^c=\mathcal{X} \setminus \mathcal{S}_i$ be its complement.
We write $U_i = U \cap \mathcal{S}_i$ for the client-supported portion of public data and $U_i^c = U\setminus U_i$ for its complement.
Let $\cQ(\mathcal{S}_i)$ denote the probability of $\mathcal{S}_i$ under $\cQ$.
Let $\cQ_i := \mathcal Q(\cdot\mid \mathcal S_i)$ denote the restriction of $\cQ$ to $\mathcal{S}_i$, with density $q_i(x) = q(x)/\cQ(\mathcal{S}_i)$.
Let $\cQ_i^c := \mathcal Q(\cdot\mid \mathcal S_i^c)$ .
For any functions $g$ and $g'$, and input $x$, define
\(
G(g,g',\ell) = \ell(g,g') + \lambda r_{\cB}(g; \ell).
\)
Then, we can write the server-side cluster $k$'s objective $\hat{J}(h_k) = \sum_{x \in U} G(h_k,\bar{f}_k,\ell)(x)$, where $h_k$ is the server-side model and $\bar{f}_k$ the aggregate model over clients in the $k$th cluster.
Similarly, we can write $\hat{F}(f_i;\ell) = \sum_{x \in U} w_i(x)G(f_i,h_{\pi(i)},\ell)(x)$ for a client $i$'s model $f_i$.
It will also be useful to define $\hat{J}_i(h_k;\ell) = \sum_{x \in U_i} G(h_k,\bar{f}_k,\ell)(x)$ and $\hat{J}_i^c(h_k;\ell) = \sum_{x \in U_i^c} G(h_k,\bar{f}_k,\ell)(x)$ for the server, and similarly, $\hat{F}_i(f_i;\ell) = \sum_{x \in U_i} w_i(x)G(f_i,h_{\pi(i)},\ell)(x)$ and $\hat{F}_i^c(f_i;\ell) = \sum_{x \in U_i^c} w_i(x)G(f_i,h_{\pi(i)},\ell)(x)$ for each client $i$.
$\hat{J}_i$ and $\hat{F}_i$ are the server and client objectives with respect to the public data $U_i$ restricted to support of $i$'s data distribution, whereas $\hat{J}_i^c$ and $\hat{F}_i^c$ are these objectives on the portion of $U$ outside client $i$'s support.
Since $U = U_i \cup U_i^c$ for each $i$, we can also note that $\hat{J}(h_k;\ell) = \hat{J}_i(h_k;\ell) + \hat{J}_i^c(h_k;\ell)$ and $\hat{F}(f_i;\ell) = \hat{F}_i(f_i;\ell) + \hat{F}_i^c(f_i;\ell)$.

Moreover, insofar as our interest is in convergence (in terms of \emph{true} risk), the objectives for the clients and server defined above are only estimates thereof.
The true objective of each server-side cluster is $J(h_k;\ell) = \mathbb{E}_{x \sim \cQ}[G(h_k,\bar{f}_k,\ell)(x)]$, with $J_i(h_k;\ell)=\cQ(\mathcal{S}_i) \mathbb{E}_{x \sim \cQ_i} [G(h_k,\bar{f}_k,\ell)(x)]; J_i^c(h_k)=\cQ(\mathcal{S}_i^c) \mathbb{E}_{x \sim \cQ_i^c} [G(h_k,\bar{f}_k,\ell)(x)]$; and we define $F(f_i;\ell)$, $F_i(f_i;\ell)$, and $F_i^c(f_i;\ell)$ analogously for each client $i$.  While cumbersome, the key consideration that this notation enables us to deal with is how well-behaved the objective functions of the server and clients are on the supported and unsupported parts of the public data.

\subsection{Clustering}
\label{S:clustering}

Our risk analysis relies on a controlled within-cluster pseudo-label distance $B$ over the shared unlabeled set $U$.
In particular, the greedy clustering procedure in Section~\ref{S:phase1} enforces this property in the first round by constructing clusters with threshold $B_0$, where $B_0 \le B$. 
In subsequent rounds, bounded in-cluster disagreement is expected to continue to hold because all clients are anchored to the same public set and distill from the same cluster-level teacher in each round. This condition is the mechanism that lets us control how much a cluster-level aggregation step can degrade client $i$'s pseudo-labels, which leads to Lemma~\ref{lem:aggregation}.
Without an explicit bound, similarity-based clustering may work well empirically but offers no convergence guarantees: aggregation within a cluster can introduce uncontrolled label error.

We additionally require clients to have confidence margin on $U_i$ to prevent small within-cluster discrepancies from flipping argmax labels.
This is reasonable because $f_i^{(t,1)}$ is obtained by fine-tuning on client $i$'s labeled data, so its predictions on the in-support public subset $U_i$ are expected to be sufficiently decisive.
Formally, for any $g(x)$ mapping inputs to a distribution over labels $[M]$, define the margin
$\Delta_g(x) = g(x)_{(1)} - g(x)_{(2)}$, where $g(x)_{(m)}$ denotes the $m$-th largest coordinate of $g(x)$.

\begin{assumption}\label{assum:cluster_margin}
For each iteration $t$ and cluster $k$, the within-cluster pseudo-label distance is bounded:
\(
\max_{i,j \in \mathcal{G}_k} d^{(t)}(i,j) \le B.
\)
Moreover, for each client $i$ there exists $\gamma>0$ such that for every $x \in U_i$,
\(
\Delta_{f_{i}^{(t,1)}}(x) \ge \gamma.
\)
\end{assumption}

\begin{lemma}\label{lem:aggregation}
For each client $i$,
\(
\err_{x\in U_i}(\bar{f}_{k}^{(t)}) \;\le\; \err_{x\in U_i}(f_{i}^{(t,1)}) \;+\; \frac{2B}{\gamma\,|U_i|}.
\)
\end{lemma}

\subsection{Data and Objective Conditions}

Our next challenge comes from the server's reliance on the
public unlabeled pool \(U\): since it never observes clients' labeled
data, any guarantee must ensure that the distribution \(Q\) induced by
\(U\) provides adequate coverage of each local distribution \(\mathcal D_i\). In particular, a key requirement is that the public distribution \(Q\) does not under-cover nor over-concentrate on regions where client $i$ has support. We formalize this as the following condition.




\begin{assumption}[Distributional coverage] \label{assum:importance}
For each client $i$ with local data distribution $\cD_i$ that has density $p_i$, $w_{i,1} \;=\;\sup_{x\in \mathcal{S}_i}\frac{p_i(x)}{q_i(x)}$ and $w_{i,2} \;=\;\sup_{x\in \mathcal{S}_i}\frac{q_i(x)}{p_i(x)}$ are bounded. 
\end{assumption}
We note that we can in principle \emph{construct} $U$ to be sufficiently diverse, or define a sampling distribution $\cQ$ to have a strictly positive density over $\mathcal{X}$, such as a Gaussian distribution.
Then, if $\mathcal{X}$ is bounded (for example, $\mathcal{X} = [0,1]^m$, as for normalized image data), the condition will hold for reasonable $\cD_i$ (since the sup is over its support).


Furthermore, we must ensure convexity for a well-conditioned optimization landscape, and that data points outside the support of each client's distribution do not excessively interfere with learning from client-supported data, where pseudo-labels are expected to be more accurate. To capture this, we require key data-dependent regularity conditions on the server and client objectives, $J, \hat{J}, F$, and $\hat{F}$, captured by the following definition.
\begin{definition}[Locally well-conditioned Objective]
We say a differentiable function $\Phi(\theta;\ell)$ is $(\mu,L)$-\emph{locally well-conditioned} if
\begin{enumerate}
  \item \(\Phi_i\) is \(\mu\)-strongly convex:
  \(
    \nabla^2 \Phi_i(\theta) \succeq \mu I\ \text{for } \mu > 0.
  \)
  \item $\nabla \Phi^c_i$ is $L_2$-Lipschitz continuous:
  \(
    \|\nabla \Phi_i^c(\theta) - \nabla \Phi_i^c(\theta')\| \le L \|\theta - \theta'\| \ \forall \theta,\theta'
  \)
  for some $0<L<\mu$.
  \item $\|\nabla \hat{\Phi}_i^c(\theta^{*})\| \le \tau$ where \(\theta^* = \arg\min_{\theta} \hat{\Phi}_i(\theta)\).
\end{enumerate}
\end{definition}

In the following, we note that the functions $F$, $\hat{F}$, $J$, and $\hat{J}$ are, effectively, functions of \emph{parameters} of the models they are constructed around, whether these are the server-side models $h_k$ or the client-side models $f_i$.
\begin{assumption}
    \label{assump:data_dep_conv_smooth}
    For all clients $i$ and COSMOS iterations $t$:
    \begin{enumerate}
    \item $F(\theta_i^{(t)};\ell)$, $\hat{F}(\theta_i^{(t)};\ell)$, $J(\theta_k^{(t)};\ell)$, and $\hat{J}(\theta_k^{(t)};\ell)$ are locally well-conditioned, and 
    \item The pseudolabels generated by any pairs of parameters $\theta, \theta'$ from the same hypothesis class are $L_1$-Lipschitz continuous:
\(
\sup_{x\in\mathcal{X}}\big\|\theta(x)-{\theta'}(x)\big\|_\infty
\;\le\; L_1\,\|\theta-\theta'\|.
\)
    
    \end{enumerate}
\end{assumption}

The final regularity condition bounds how often inputs fall arbitrarily close to the decision boundary, so that a small change in the objective does not frequently flip the predicted label. Concretely, we adopt a standard Tsybakov-style margin condition for the optimal classifiers.

\begin{assumption}[Confidence margin condition]\label{assum:margin}

For each client $i$, let $f_i^* \in \arg\min_{g \in \mathcal{H}_i} F_i(g;\ell)$ and let $h_i^* \in \arg\min_{g \in \mathcal{H}_S} J_i(g;\ell)$. Then for all $i$, there exist constants $C>0$ and $\alpha>0$ such that $\forall t\ge 0$,
\(
\Pr\big(\Delta_{f_i^*}(x) \le t\big) \;\le\; C\, t^{\alpha}
\)
and
\(
\Pr\big(\Delta_{h_i^*}(x) \le t\big) \;\le\; C\, t^{\alpha}.
\)
\end{assumption}

\subsection{Pseudolabel and Label Conditions} \label{subsec:expansion}
While our clustering controls the additional error introduced by aggregation, and our data/objective conditions support stable gradient-based optimization, a purely pseudo-label-based method still requires additional structure to be provably effective. In particular, the pseudo-labels must carry nontrivial information about the ground truth, and the ground truth labels should satisfy some structural connectivity so that the learner can generalize beyond the pseudo-labeled points rather than merely memorizing arbitrary functions. These conditions are what enable learning from pseudo-labels, inducing the weak-to-strong generalization we observe at the server's end~\cite{lang2024theoretical}.

Specifically, we adopt the expansion-and-robustness framework of \cite{Wei2020TheoreticalAO}, which expresses label connectivity via input-space transformations $\mathcal{T}$ and associated neighborhoods $\mathcal{N}(x)$ of a given input $x$ (see Section~\ref{S:phase1}, Step 3).
For a subset $V \subseteq \mathcal X$, define $\mathcal N(V)$ as the union of neighborhoods of its points:
$\mathcal N(V) = \bigcup_{x \in V} \mathcal N(x)$.
Expansion ensures that if a set $V$ contains a small fraction of a given class, then closing $V$ under the neighborhood operator $\mathcal{N}(\cdot)$ captures a larger fraction of that same class.

\begin{definition}[$(b,c)$-expansion] A distribution $\cD$ satisfies $(b,c)$-expansion if for every class $m$ and $\forall V \subseteq\mathcal{X}$ with $\Pr_{x \sim \cD|x \in V}[y(x) = m] \le b$, we have $\Pr_{x \sim \cD|x\in \cN(V)}[y(x) =m] \ge \min\{c\Pr_{x \sim \cD}[y(x) = m \ |\ c\in V],1\}$.
\end{definition}

To quantify the probability that pseudo-labelers make mistakes for any client $i$ and iteration $t$, define
\begin{align*}
\overline{b}
=\sup_{t,i}\max\Bigl\{
\cQ_i\!\left(\mathcal{M}(h_{\pi(i)}^{(t)})\right),\;
\cQ_i\!\left(\mathcal{M}(\bar{f}_{\pi(i)}^{(t)})\right),\;\\[3pt]
\cD_i\!\left(\mathcal{M}(h_{\pi(i)}^{(t)})\right),\;
\cD_i\!\left(\mathcal{M}(\bar{f}_{\pi(i)}^{(t)})\right)
\Bigr\},
\end{align*}
where \(\mathcal M(f) := \{x \in \mathcal X : A \circ f(x) \ne y(x)\}\).

The following natural condition requires that the probability $\overline{b}$ of pseudo-labeling errors is not too large, as well as inputs with a given class are sufficiently well-connected.

\begin{assumption}[Effective pseudo-labelers]\label{assump:expansion}
$\overline{b} \le \tfrac{1}{3}$.

\end{assumption}

Complementing expansion, robustness formalizes the idea that inputs close in the input space share the same labels. We characterize this property in terms of robustness loss.
\begin{definition}
For a function $f$ and distribution $\cD$,  \emph{robustness loss} is the fraction of examples that are not robust to input transformations:
\[
R_{\mathcal{B}}(f,\cD)
\;=\;
\mathbb{E}_{x\sim \mathcal{D}}\Bigl[\mathbf{1}\bigl\{\exists\,x'\in \mathcal N(x)\colon f(x')\neq f(x)\bigr\}\Bigr].
\]
\end{definition}
\begin{assumption}[Label connectivity] \label{assum:robustness} For every client $i$ and true classifier $y(x)$,
\(\mathcal{Q}_i\) and \(\mathcal{D}_i\) satisfy \((\overline{b}, \overline{c})\)-expansion for some \(\overline{c}>3\) and 
$\max\{R_{\cB}(y,\cQ_i),$ $R_{\cB}(y,\cD_i)\}\le\;\rho$.
\end{assumption}

\subsection{Risk Contraction}
We now put these tools together to provide bounds for the client-side and server-side risk.
\ifwithappendix
The proofs are provided in the Appendix (\ref{S:proofs_risk_bounds}).
\else
Proofs are omitted for space.
\fi
Define $c = \min\left\{\overline{c},\,\frac{1}{\overline{b}}\right\}$. 
Suppose that
1) $c > 2 w_1 w_2 + 1,$ where $w_1=\max_i\{w_{i,1}\}$ and $w_2=\max_i\{w_{i,2}\}$, 
2) $\lambda = \tfrac{2c}{c+1}$, 3) Conditions~\ref{assum:cluster_margin}--\ref{assum:robustness} hold, and 4) each supervised local update is classification-safe: with high probability, the update from $f_i^{(t-1,2)}$ to $f_i^{(t,1)}$ does not increase the true $0$--$1$ risk on $\cD_i$ beyond the lower-order generalization slack.

\begin{theorem}\label{thm:server_improvement}
    For any client $i$ at any iteration $t$, with probability at least $1 - \delta$, its server model $h_
{\pi(i)}^{(t)}$'s risk is contracting:
\(
  R_{\cD_i}(h_
{\pi(i)}^{(t)}) \;\le\; \kappa_1\, R_{\cD_i}(f_i^{(t,1)}) + \tilde{O}(n^{-1/2}),
\)
where $\kappa_1=\tfrac{2w_1w_2}{c-1}<1$ and the lower order term hides constants and poly-log terms in $n$.
\end{theorem}

\begin{theorem}\label{thm:client_convergence}
    Let $\kappa_2=\tfrac{4 w_1 w_2}{(c-1)^2}<1$.
For each client \(i\) and iteration $t$,  with probability at least \(1-\delta\),
\(
R_{\mathcal{D}_i}\!\left(f_i^{(t,2)}\right)
\;\le\;
\kappa_2\,R_{\mathcal{D}_i}\!\left(f_i^{(t-1,2)}\right)
+\;
\tilde{O}\!\left(\sqrt{\frac{\log(1/\delta)}{n}}\right).
\)

\end{theorem}

Applying a union bound, we obtain a contraction of the personalization risk bound at an exponential rate.
\begin{corollary}[Personalization Risk Bound]\label{cor:personalization}
With probability at least 
$1-\delta$
personalization risk after $T$ iterations is
\begin{equation*}
R^\dagger\!\left(h^{(T)}_{1},\ldots,h^{(T)}_{K}\right)
\le \frac{1}{N}\sum_{i=1}^{N}\kappa_2^{\,T} R_{\mathcal{D}_i}\!\left(f_i^{(1,1)}\right)
+ \tilde{O}\!\left(\sqrt{\frac{\log(T/\delta)}{n}}\right).
\end{equation*}
\end{corollary}
Moreover, for sufficiently large $n$ and $T$, personalization risk drops below the risk from local training alone.

\setlength{\tabcolsep}{5pt}
\renewcommand{\arraystretch}{1.1}

\begin{figure*}[t]
\centering
\small
\includegraphics[width=\textwidth]{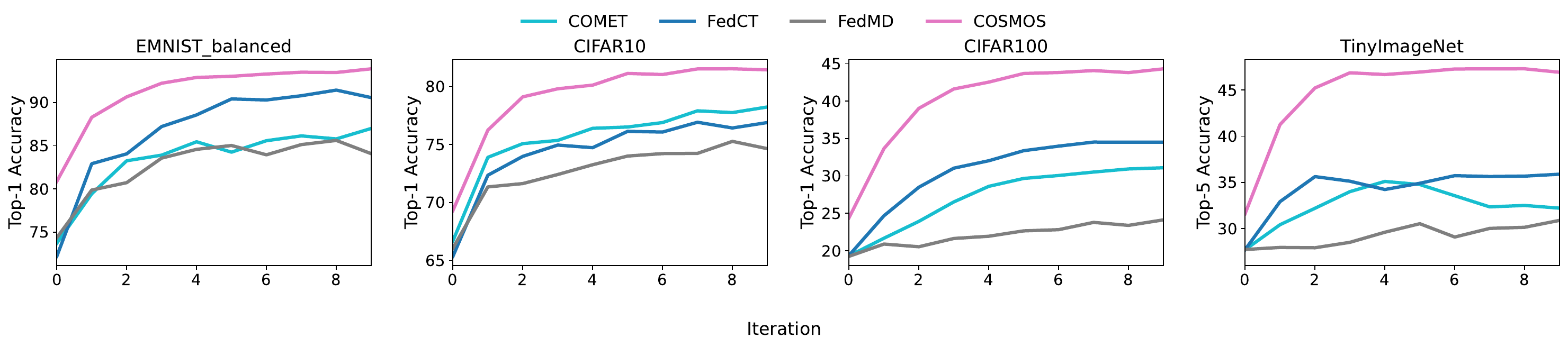}
\vspace{-0.5em}
\caption{Comparison of COSMOS and model-agnostic baselines on four benchmarks. 
Client models consist of MobileNet and SqueezeNet architectures. Curves show mean client accuracy over rounds (Top-1 for  CIFAR-10/100 and EMNIST; Top-5 for Tiny ImageNet).\ifwithappendix\ A statistical comparison of final-round performance is provided in the Appendix (Appendix~\ref{appendix:additional_results}, Table~\ref{tab:benchmarks-final}).\fi}

\label{fig:diff_benchmarks_05}

\end{figure*}

\begin{figure*}[t]
\centering

\includegraphics[width=\textwidth]{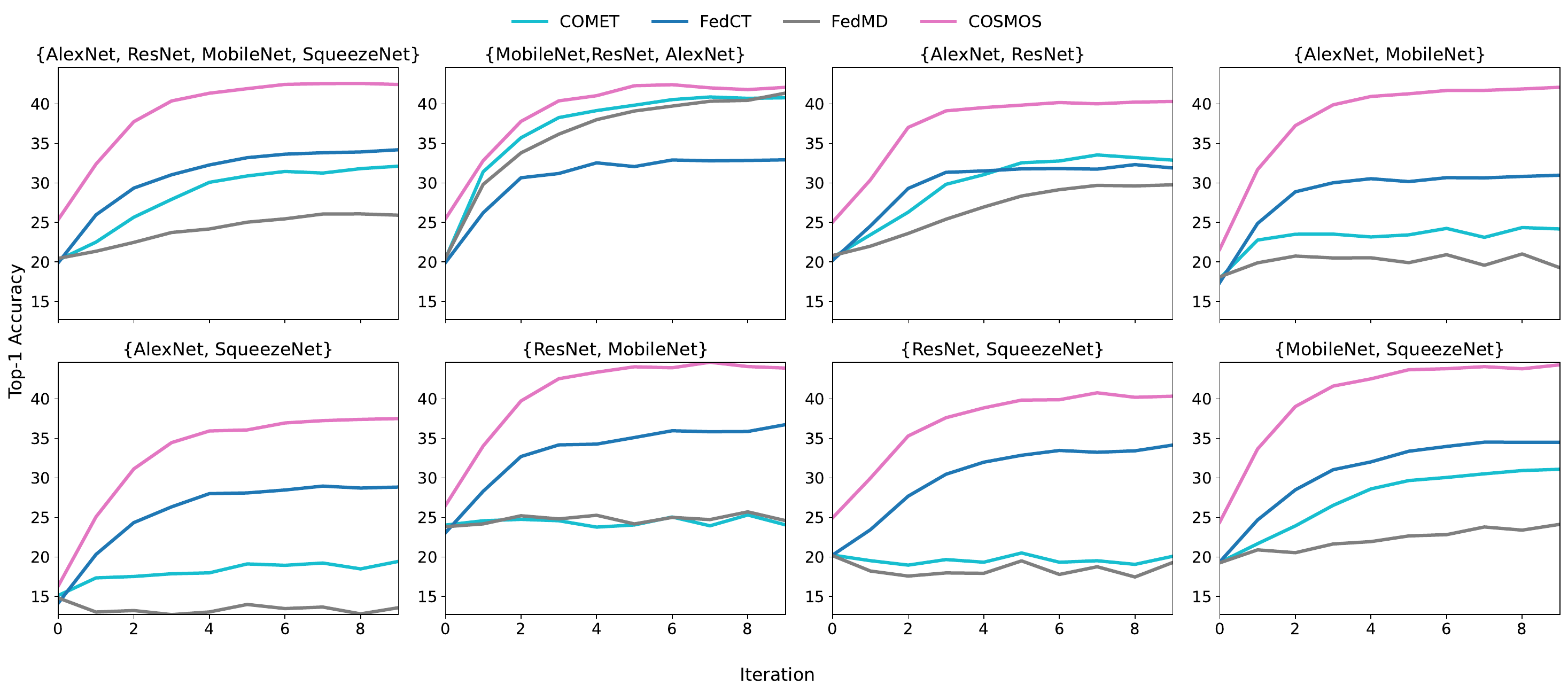}

\caption{Performance of COSMOS and heterogeneous-model baselines on CIFAR-100 under different mixtures of client architectures. Each subfigure corresponds to a distinct architecture combination.\ifwithappendix\ A statistical summary of the final communication round appears in Appendix~\ref{appendix:additional_results}, Table~\ref{tab:arch-mixtures-final}.\fi
 }
\label{fig:diff_clients_nets}

\end{figure*}

\section{Experiments}\label{S:exp}

\noindent\textbf{Experiment Setup:}
We construct client datasets using Dirichlet non-IID partitioning with concentration parameter $\alpha$~\cite{yurochkin2019bayesian}. Specifically, the label space is partitioned into five disjoint groups, each containing $20\%$ of the classes, and clients are assigned to one group. For each class, its samples are distributed among the clients within the same group according to a Dirichlet draw. 
To introduce limited cross-group exposure and better reflect realistic data sharing, a fraction of each client's local samples (10\%) is pooled, randomly shuffled, and uniformly redistributed across all clients.
Our unlabeled dataset $U$ consists of 20\% of all training data.
We use accuracy as the efficacy measure. 

We selected client architectures among 4 options: AlexNet, ResNet18, MobileNet, and SqueezeNet.
We use VGG16 for the server.
Our focus is on the model-agnostic setup, where we compare COSMOS with the only three truly model-agnostic baselines (all others require knowledge of client model architecture; see Section~\ref{S:relwork}): FedMD~\cite{li2019fedmd}, FedCT~\cite{abourayya2025little}, and COMET~\cite{cho2023communication}.
\ifwithappendix
In the Appendix, we also evaluate COSMOS in \emph{homogeneous model} settings, where we compare to \emph{eight state-of-the-art FL baselines}, focusing primarily on PFL approaches.
\fi

We use four image classification benchmarks: EMNIST-balanced~\cite{cohen2017emnist}, CIFAR-10 and CIFAR-100~\cite{krizhevsky2009learning}, and Tiny ImageNet~\cite{le2015tiny}. 
The main experiments use 25 clients over 10 communication rounds, using Dirichlet non-IID sampling ($\alpha = 5$). 
Each data point in the figures reflects the \emph{mean client accuracy} over three random seeds.
\ifwithappendix
We present further details about the \emph{experiment setup, confidence intervals, scalability evaluation of COSMOS as we increase the number of clients, and other parametric ablations, in the Appendix.}
\fi
Because all compared model-agnostic baselines exchange predictions on the shared pool, communication cost is naturally tied to the size of $U$ and the label dimension; even so, prediction exchange remains far cheaper than parameter sharing in our setting. For example, the per-client transmission cost is $0.38$ MB for pseudo-labels versus $113.36$ MB for parameter transfer on CIFAR-10, and $3.81$ MB versus $114.76$ MB on CIFAR-100.



\smallskip
\noindent\textbf{Results:}
Figure~\ref{fig:diff_benchmarks_05} presents the main results across all four benchmarks in the model-agnostic setting with $\alpha=5$. 
In all cases, COSMOS outperforms all model-agnostic baselines by a large margin.
Moreover, its advantage increases with increasing task complexity (from EMNIST to TinyImageNet).
This trend is stable across the additional heterogeneity settings $\alpha \in \{1,100\}$\ifwithappendix\ reported in the Appendix\fi: COSMOS remains strongest on the harder benchmarks and does not lose its advantage when heterogeneity is either increased or relaxed.\ifwithappendix\ Confidence-bound information is also reported in the Appendix.\fi


Figure~\ref{fig:diff_clients_nets} evaluates COSMOS under different mixtures of client architectures. 
Across all combinations, COSMOS remains robust to heterogeneity in client capabilities, including settings with a high proportion of weaker models (AlexNet), maintaining substantial performance edge over baselines. 
\ifwithappendix
Additionally, our evaluation in the homogeneous architecture setting (see the Appendix, Appendix~\ref{appendix:additional_results}, Figure~\ref{fig:homog_arch}) demonstrates that COSMOS remains competitive in such environments with the best-performing PFL and model-agnostic baselines.
This is particularly notable, as the best-performing baseline \emph{changes for different model architectures}, further demonstrating the value of our model-agnostic design.
Moreover, our analysis of the communication cost demonstrates the bandwidth advantage of COSMOS over prior art (see the Appendix, Table~\ref{table:communication_comparison}).
The same appendix also makes clear that this benefit is not free: performance improves with both the size of the shared public pool and the capacity of the server model, which is consistent with the theoretical coverage assumptions and with the additional computation required to train cluster-specific teachers.
\else
These gains are not free: COSMOS improves further when the shared public pool is larger and the server model is stronger, which is consistent with the theoretical coverage assumptions and with the additional computation required to train cluster-specific teachers.
\fi


Finally, we analyzed the sensitivity of COSMOS to key design choices and hyperparameters (including temperature $T$ and weight of the regularization term $\lambda$). 
We find that COSMOS is quite robust to small changes in $\lambda$, with $T=1$ and $\lambda=5$ yielding the best performance.
\ifwithappendix
Additionally, we observe improvement in COSMOS performance as the number of clients or the size of $U$ increase, as well as with the use of a more powerful server model architecture; see the Appendix for further details.
\fi

\section{Conclusion}
In this work, we introduce COSMOS, a framework that addresses a critical bottleneck in federated learning: achieving high-performance personalization while remaining fully model-agnostic. By allowing clients to utilize arbitrary (even proprietary) architectures through the communication of predictions on a shared unlabeled pool, COSMOS removes the structural barriers that have historically limited the deployment of PFL in diverse, real-world ecosystems.
We establish a general end-to-end theoretical analysis of risk (generalization) bound contraction of our framework that significantly generalizes past theoretical results in this setting. We also validate the algorithm's effectiveness against state-of-the-art baselines across multiple benchmarks. Current experiments are limited to image classification, and a key next step is to study more severe public/private distribution shift as well as additional modalities and decentralized variants of the framework.

\section*{Acknowledgements}
This work was supported in part by the National Science Foundation (IIS-2214141, CCF-2403758), Army Research Office (W911NF-25-1-0059), and Office of Naval Research (N00014-24-1-2663).

\bibliographystyle{splncs04}
\bibliography{mybibliography}

\clearpage
\appendix
\section{Notation Table}\label{app:notation}%

\begin{table}[h!]
\caption{Summary of the main notation used throughout this paper.}
\label{table:notation}
\centering
\begin{tabular}{l p{6.5cm}}
        \toprule
        \textbf{Notation} & \textbf{Description} \\
        \midrule
        $N,i$ & Number of clients, client index \\
        $K,k$ & Number of clusters, cluster index\\ 
        $\mathcal{G}_k$ & Set of clients in cluster $k$ \\
        $\pi: [N] \rightarrow [K]$ & Assignment function of clients. \\
        $\mathcal{X}, M$ & Domains of inputs, number of classes \\
        $y(\cdot) \in [M]$ & Ground-truth label \\
        $D_i, \mathcal{D}_i, p_i(\cdot)$ & Client $i$'s labeled dataset, its distribution, and density function \\
        $U, \mathcal{Q}, q(\cdot)$ & Global dataset,  distribution, and density\\
        $S_i $ & Support of client $i$'s distribution. \\
        $U_i $ &  $U$ restricted to $S_i$ \\
        $\mathcal{Q}_i$ & $\mathcal{Q}$ restricted to $S_i$ \\

       \midrule

        $f_i^{(t,1)}$ & Client $i$'s model at iteration $t$ after training on local data \\
        $f_i^{(t,2)}$ & Client $i$'s model at iteration $t$ after training on the server's feedback \\
        $\bar{f}^{(t)}_k=\frac{1}{|\mathcal{G}_k|}\sum_{i \in \mathcal{G}_k} f_{i}^{(1,1)}$ & Aggregator of cluster $k$ at iteration $t$ \\
        $h_k^{(t)}$ & Server model for cluster $k$ at iteration $t$ \\
    
        $R_{\mathcal{D}}(g)$ & Classification risk of a model $g$ under $\mathcal{D}$ \\
        $\err_{x\in D}(g)$ & Average classification error of a model $g$ over a finite dataset $D$ \\
        $R^\dagger(h_1,\dots,h_K)$ & Server's personalization risk \\
        
        $\hat{J}(h_k;\ell)= \hat{J}_i+ \hat{J}_{i}^{c}$ & Cluster $k$'s empirical training objective partitioned according to client $i$'s support \\
        $J_k(\cdot)$ & Cluster $k$'s population training objective \\
        
        $\hat{F}(f_i;\ell) = \hat{F}_{i}(f_i;\ell)+ \hat{F}_{i}^c(f_i;\ell)$ & Client $i$'s empirical training objective. \\
        $F_i(\cdot)$ & Client $i$'s population training objective \\
        $\mathcal{B}(x), \mathcal{N}(x)$ & Transformation ball and neighborhood of example $x$ defined by $\mathcal{T}(x)$. \\
        
        $r_\mathcal{B}(g;\ell)(x) $ & Regularization loss for $S\subset \mathcal{N}(x)$\\
        \midrule
        $B_0,B$ & Hyperparameter for the greedy clustering step, universal bound over in-cluster distance\\
        $\Delta_g(x)$ &Confidence margin of model $g$ over point $x$ \\
        $w_1,w_2$ & Universal bounds over the importance weightings between $\mathcal{Q}_i$ and $\mathcal{D}_i$. \\
        $\bar{b},\bar{c},\rho$ & Expansion and robustness constants \\
        $\kappa_1, \kappa_2$ & Risk bounds' contraction coefficients \\
        \bottomrule
\end{tabular}
\end{table}

\section{Omitted Proofs}

\subsection{Proof of Lemma~\ref{lem:aggregation}} \label{S:clustering_proof}

\begin{proof}

Let $\hat c_i(x):=\arg\max_{c\in[M]} f_i^{(t,1)}(x)_c$ and $\hat c_k(x):=\arg\max_{c\in[M]} \bar f_k^{(t)}(x)_c$.
For any $x\in U_i$ we have the pointwise implication
\[
\mathbf{1}\{\hat c_k(x)\neq y(x)\}
\;\le\;
\mathbf{1}\{\hat c_i(x)\neq y(x)\}+\mathbf{1}\{\hat c_k(x)\neq \hat c_i(x)\},
\]
because if $\hat c_k(x)=\hat c_i(x)$ then $\hat c_k(x)\neq y(x)$ implies $\hat c_i(x)\neq y(x)$.
Now consider the set of public supported points where
the averaged predictor disagrees with client $i$ on the hard classification label. Call it $\mathrm{Bad}:=\{x\in U_i:\hat c_k(x)\neq \hat c_i(x)\}$. 

Averaging over $U_i$ gives
\begin{equation}
\err_{x\in U_i}\!\bigl(\bar f_k^{(t)}\bigr)
\;\le\;
\err_{x\in U_i}\!\bigl(f_i^{(t,1)}\bigr)
+\frac{1}{|U_i|}\,|\mathrm{Bad}|.
\label{eq:err_cluster}
\end{equation}

It remains to bound $|\mathrm{Bad}|$.
Fix any $x\in \mathrm{Bad}$, and denote $c^\star:=\hat c_i(x)$ and $\tilde c:=\hat c_k(x)$.
By definition of $\mathrm{Bad}$ we have $\tilde c\neq c^\star$ and
\[
\bar f_k^{(t)}(x)_{\tilde c}\;\ge\;\bar f_k^{(t)}(x)_{c^\star}.
\]
Define the coordinate deviations $\delta_c := \bar f_k^{(t)}(x)_c - f_i^{(t,1)}(x)_c$.
Rearranging the inequality above yields
\( \delta_{\tilde c}-\delta_{c^\star}
\;\ge\;
f_i^{(t,1)}(x)_{c^\star}-f_i^{(t,1)}(x)_{\tilde c}.
\)
By the margin assumption (applied with competitor class $\tilde c$),
$f_i^{(t,1)}(x)_{c^\star}-f_i^{(t,1)}(x)_{\tilde c}\ge \gamma$,
hence
\(
\delta_{\tilde c}-\delta_{c^\star}\;\ge\;\gamma.
\)
Therefore, by the triangle inequality,
\begin{align*}
    \bigl\|\bar f_k^{(t)}(x)-f_i^{(t,1)}(x)\bigr\|_1 
=&\sum_{c\in[M]}|\delta_c|
\;\\
\ge&\;|\delta_{\tilde c}|+|\delta_{c^\star}|
\;\\
\ge&\;|\delta_{\tilde c}-\delta_{c^\star}|
\;\ge\;\gamma.
\end{align*}

Summing this lower bound over $x\in\mathrm{Bad}$ gives
\begin{align*}
    \label{eq:bad-lower}
&\sum_{x\in U_i}\bigl\|\bar f_k^{(t)}(x)-f_i^{(t,1)}(x)\bigr\|_1\\
&\;\ge\;
\sum_{x\in \mathrm{Bad}}\bigl\|\bar f_k^{(t)}(x)-f_i^{(t,1)}(x)\bigr\|_1\\
&\;\ge\;
\gamma|\mathrm{Bad}|.
\end{align*}

On the other hand, let us denote client $i$'s correspond cluster as $|\mathcal G_k|$.

Using $\bar f_k^{(t)}(x)-f_i^{(t,1)}(x)=\frac{1}{|\mathcal G_k|}\sum_{j\in \mathcal G_k}\bigl(f_j^{(t,1)}(x)-f_i^{(t,1)}(x)\bigr)$ and the triangle inequality, we get
\begin{align*}
    \bigl\|\bar f_k^{(t)}(x)-f_i^{(t,1)}(x)\bigr\|_1
\;\le\;
\frac{1}{|\mathcal G_k|}\sum_{j\in \mathcal G_k}\bigl\|f_j^{(t,1)}(x)-f_i^{(t,1)}(x)\bigr\|_1.
\end{align*}

As the greedy clustering algorithm proceeds by putting all the clients that are $B$-close to the center into one cluster, two clients from the same cluster's distance is bounded by $2B$. Summing over $x\in U_i$ yields
\begin{align*}
&\sum_{x\in U_i}\bigl\|\bar f_k^{(t)}(x)-f_i^{(t,1)}(x)\bigr\|_1\\
&\le
\frac{1}{|\mathcal G_k|}\sum_{j\in \mathcal G_k}\sum_{x\in U_i}\bigl\|f_j^{(t,1)}(x)-f_i^{(t,1)}(x)\bigr\|_1\\
&\le
\frac{1}{|\mathcal G_k|}\sum_{j\in \mathcal G_k} 2B
\;=\;2B.
\end{align*}
Thus $|\mathrm{Bad}|\le 2B/\gamma$.
Plugging into Equation~(\ref{eq:err_cluster}) yields
\[
\err_{x\in U_i}\!\bigl(\bar f_k^{(t)}\bigr)
\;\le\;
\err_{x\in U_i}\!\bigl(f_i^{(t,1)}\bigr)
+\frac{1}{|U_i|}\cdot\frac{2B}{\gamma},
\]
which establishes the desired bound.
\end{proof}

\subsection{Proofs of Risk Bounds}
\label{S:proofs_risk_bounds}

To prove our main results, which are the clients' and server's risk bounds, we begin with the following key lemma.

\begin{lemma}[Stability under objective shift]
\label{lem:objective_shift_stability} Fix an iteration $t$, and a client $i$, let \(\Phi\) denote either the client-side objective \( F_i \) or the server-side objective \( J_i \) in the surrogate version. Let \(\hat{\theta}\) represent the corresponding model minimizing $\hat{\Phi}$ (i.e., \( f_i^{(t,2)} \) or \( h_{\pi(i)}^{(t)} \)), and $\theta$ represent $\arg\min_\Theta \Phi_i(\theta)$ then 
\[
  R_{\cQ_i}(\hat{\theta}) - R_{\cQ_i}(\theta^*) 
  \;\le\; C\left(2L_1 (\frac{\tau}{\mu - L_2} + \sqrt{\frac{4\varepsilon_n}{\mu}})\right)^\alpha.
\]
and
\[
  R_{i}(\hat{\theta}) - R_{i}(\theta^*) 
  \;\le\; C\left(2L_1 (\frac{\tau}{\mu - L_2} + \sqrt{\frac{4\varepsilon_n}{\mu}})\right)^\alpha.
\]

\end{lemma}

\begin{proof}
Recall the empirical objective can be decomposed into supported and unsupported parts:
\(
\hat{\Phi}(\theta) = \hat{\Phi}_{i}(\theta) + \hat{\Phi}_{i}^c(\theta)
\).
Let $\hat{\theta}_1 := \arg\min_\Theta \hat{\Phi}_i(\theta)$.  Since $\hat{\theta}$ minimizes $\hat{\Phi}$, we have $\nabla \hat{\Phi}_i(\hat{\theta}) + \nabla \hat{\Phi}_i^c(\hat{\theta}) = 0$. Using the mean value theorem on $\nabla \hat{\Phi}_i$ between $\hat{\theta}$ and $\hat{\theta}_1$:
\[
\nabla \hat{\Phi}_i(\hat{\theta}) = \nabla^2 \hat{\Phi}_i(\bar{\theta})(\hat{\theta} - \hat{\theta}_1),
\]
for some $\bar{\theta}$ on the line segment between $\hat{\theta}_1$ and $\hat{\theta}$.

Because $\nabla \hat{\Phi}_i(\hat{\theta}_1) = 0$ and $\hat{\Phi}_i$ is $\mu$-strongly convex, we obtain
\begin{align*}
\|\hat{\theta} - \hat{\theta}_1\|
&= \|\nabla^2 \hat{\Phi}_i(\bar{\theta})^{-1} \nabla \hat{\Phi}_i(\hat{\theta})\| \\
&\le \frac{1}{\mu} \|\nabla \hat{\Phi}_i(\hat{\theta})\| = \frac{1}{\mu} \|\nabla \hat{\Phi}_i^c(\hat{\theta})\|.
\end{align*}

Using $L_2$-smoothness of $\hat{\Phi}_i^c$, we bound
\[
\|\nabla \hat{\Phi}_i^c(\hat{\theta})\| \le \|\nabla \hat{\Phi}_i^c(\hat{\theta}_1)\| + L_2 \|\hat{\theta} - \hat{\theta}_1\|.
\]
Rearranging,
\[
\|\hat{\theta} - \hat{\theta}_1\| \le \frac{1}{\mu - L_2} \|\nabla \hat{\Phi}_i^c(\hat{\theta}_1)\| \le \frac{\tau}{\mu - L_2}.
\]

We now relate $\hat{\theta}_1$ to $\theta^*$. By standard uniform convergence arguments (e.g., Rademacher complexity), with high probability $1-\delta$ over the training sample,
\[
\sup_{\theta} \left| \hat{\Phi}_i(\theta) - \Phi_i(\theta) \right| \le \varepsilon_n, \quad \text{where } \varepsilon_n \in \tilde{O}(\sqrt{\frac{\log(1/\delta)}{n}}).
\]
Using this, and the fact that $\hat{\theta}_1$ minimizes $\hat{\Phi}_i$, we get:
\[
\Phi_i(\hat{\theta}_1) \le \hat{\Phi}_i(\hat{\theta}_1) + \varepsilon_n \le \hat{\Phi}_i(\theta^*) + \varepsilon_n \le \Phi_i(\theta^*) + 2\varepsilon_n.
\]
By strong convexity of $\Phi_i$, we obtain
\[
\|\hat{\theta}_1 - \theta^*\| \le \sqrt{\frac{4\varepsilon_n}{\mu}}.
\] Then by triangle inequality:
\[
\|\hat{\theta} - \theta^*\| \le \|\hat{\theta} - \hat{\theta}_1\| + \|\hat{\theta}_1 - \theta^*\| \le \frac{\tau}{\mu - L_2} + \sqrt{\frac{4\varepsilon_n}{\mu}}.
\]

Now consider the pseudolabels generated by $\hat{\theta}$ and $\theta^*$ as $\hat s, s$ respectively. Let $c^\star$ be the unique maximizer of $s$. If $\|\hat s-s\|_\infty\le \varepsilon$, then
$|\hat s_j - s_j|\le \varepsilon$ for all $j\in[M]$. In particular,
$\hat s_{c^\star} \ge s_{c^\star}-\varepsilon$ and, for any $j\neq c^\star$, $\hat s_j \le s_j+\varepsilon$.
Hence for each $j\neq c^\star$,
\[
\hat s_{c^\star}-\hat s_j \;\ge\; (s_{c^\star}-\varepsilon) - (s_j+\varepsilon)
\;=\; (s_{c^\star}-s_j) - 2\varepsilon.
\]

We can rank the $M$ scores for input $x$ by $s$ as $s_{(1)} >s_{(2)} > \dots> s_{(M)}.$ Since $s_j \le s_{(2)}$ for all $j\neq c^\star$ and $s_{c^\star}=s_{(1)}$, we have
$s_{c^\star}-s_j \ge s_{(1)}-s_{(2)} = \Delta(s)$, and thus
\[
\hat s_{c^\star}-\hat s_j \;\ge\; \Delta(s) - 2\varepsilon.
\]
If $\Delta(s) > 2\varepsilon$, then $\hat s_{c^\star}-\hat s_j>0$ for all $j\neq c^\star$,
so $c^\star$ remains the unique maximizer of $\hat s$.

Then applying the condition \[
\sup_{x\in\mathcal{X}}\big\|s_{\hat{\theta}}(x)-s_{\theta
^*}(x)\big\|_\infty
\;\le\; L_1\,\|\hat{\theta}-\theta^*\|.
\] to the bound $\varepsilon$, 
we have

\begin{align*}
  |R(\hat\theta)-R(\theta^\star)|\;\le &\; \mathbb{P}\big(h_{\hat\theta}(x)\neq h_{\theta^\star}(x)\big)\\
\;\le 
&\;
\mathbb{P}\!\left(\Delta(s) \le 2L_1 (\frac{\tau}{\mu - L_2} + \sqrt{\frac{4\varepsilon_n}{\mu}})\right)\\
\;\le&\;
C\left(2L_1 (\frac{\tau}{\mu - L_2} + \sqrt{\frac{4\varepsilon_n}{\mu}})\right)^\alpha.  
\end{align*}

\end{proof}

Next we prove the risk bounds, restating the theorems for completeness.

\begin{proof} [of Thm.~\ref{thm:server_improvement}]
Since $Q_i$ and $D_i$ has bounded importance weights, \[
1 \;=\; \int_{S_i} p_i(x)\,dx
\;\le\; \sup_{x\in S_i}\frac{p_i(x)}{q(x)} \int_{S_i} q(x)\,dx
\;=\; w_{i,1}\,Q(S_i),
\]
so $Q(S_i)\ge 1/w_{i,1}>0$.
Applying a multiplicative Chernoff bound on data drawn i.i.d. from $Q$, we have $|U_i| \in O( Q(S_i) n)= O(n)$ with probability at least $1- \exp(-n/8w_{i,1})$. And Lemma~\ref{lem:aggregation}  the aggregated pseudolabel model $\bar{f}_{k}^{(t)}$ yields:
\[
  \err_{U_i}(\bar{f}_k^{(t)}) \;\le\; \err_{{U_i}}(f_i^{(t,1)}) + \varepsilon_1, \text{where } \varepsilon_1 \in \tilde{O}(Bn^{-1}).
\]

A standard Rademacher complexity bound then gives, with high probability:
\[
  R_{\mathcal{Q}_i}(\bar{f}_k^{(t)}) 
  \;\le\; R_{\mathcal{Q}_i}(f_i^{(t,1)}) + \varepsilon_1 + \varepsilon_2, \quad \text{where } \varepsilon_2 \in \tilde{O}(n^{-1/2}).
\]

Let $h^*$ be the optimizer for $J_i$. Applying Theorem 4.3 of~\cite{Wei2020TheoreticalAO}, we obtain:
\begin{align*}
  R_{\mathcal{Q}_i}(h^*)
  &\le \frac{2}{c-1} \, R_{\cQ_i}(\bar{f}_k^{(t)}) + \frac{2c\rho}{c-1} \\
  &\le \frac{2}{c-1} \left(  R_{\mathcal{Q}_i}(f_i^{(t,1)}) + \varepsilon_1 + \varepsilon_2 \right) + \frac{2c\rho}{c-1}.
\end{align*}

Using the importance weighting trick and bounded density ratios, we have:

\begin{align*}
R_i\!\left(h_k^{(t)}\right) 
&= \mathbb{E}_{x \sim \mathcal{D}_i}\!\left[
    \ell_{0\text{-}1}\!\left(h_k^{(t)}(x),\, y(x)\right)
\right] \\
&= \mathbb{E}_{x \sim \mathcal{Q}_i}\!\left[
    \tfrac{p(x)}{q(x)}\,
    \ell_{0\text{-}1}\!\left(h_k^{(t)}(x),\, y(x)\right)
\right] \\
&\le w_1\, R_{\mathcal{Q}_i}\!\left(h_k^{(t)}\right) \end{align*}

By Lemma~\ref{lem:objective_shift_stability},
\begin{align*}
    &R_i\!\left(h_k^{(t)}\right) 
    \le w_1 \biggl(
      R_{\mathcal{Q}_i}(h^*)
      + C\Bigl(
          2L_1 \Bigl(
              \tfrac{\tau}{\mu - L_2}
              + \sqrt{\tfrac{4\varepsilon_n}{\mu}}
          \Bigr)
        \Bigr)^\alpha
    \biggr) \\
&\le w_1 \left(
      \tfrac{2}{c-1}\, R_{\mathcal{Q}_i}\!\left(f_i^{(t,1)}\right)
      + \tfrac{2(\varepsilon_1+\varepsilon_2+c\gamma)}{c-1}
    \right) \\
&\qquad
    + w_1\,
      C\Bigl(
        2L_1 \Bigl(
            \tfrac{\tau}{\mu - L_2}
            + \sqrt{\tfrac{4\varepsilon_n}{\mu}}
        \Bigr)
      \Bigr)^\alpha \\
&= \tfrac{2w_1}{c-1}\,
   \mathbb{E}_{x \sim \mathcal{D}_i}\!\left[
      \tfrac{q(x)}{p(x)}\,
      \ell\!\left(f_i^{(t,1)}(x),\, y(x)\right)
   \right] \\
&
   + \tfrac{2w_1(\varepsilon_1+\varepsilon_2+c\gamma)}{c-1}
   + w_1\,
     C\Bigl(
        2L_1 \Bigl(
            \tfrac{\tau}{\mu - L_2}
            + \sqrt{\tfrac{4\varepsilon_n}{\mu}}
        \Bigr)
     \Bigr)^\alpha \\
&\le \tfrac{2w_1 w_2}{c-1}\,
    R_i\!\left(f_i^{(t,1)}\right)
   + \tfrac{2w_1(\varepsilon_1+\varepsilon_2+c\gamma)}{c-1} \\
&
   + w_1\,
     C\Bigl(
        2L_1 \Bigl(
            \tfrac{\tau}{\mu - L_2}
            + \sqrt{\tfrac{4\varepsilon_n}{\mu}}
        \Bigr)
     \Bigr)^\alpha .
\end{align*}

By assumption, $\kappa_1:=\tfrac{2w_1w_2}{c-1}<1$ and the claim follows.
\end{proof}

\begin{proof} [of Thm.~\ref{thm:client_convergence}]

By Lemma~\ref{lem:objective_shift_stability}, with probability at least $1 - \delta$, the difference of the true risk between the obtained model, and the minimizer $f^*$ for $\mathcal{F}_{i,\mathrm{supp}}$ is bounded as:
\[
  R_{i}(f_
{i}^{(t,2)}) - R_{i}(f^*) 
  \;\le\; C\left(2L_1 (\frac{\tau}{\mu - L_2} + \sqrt{\frac{4\varepsilon_n}{\mu}})\right)^\alpha.
\]

Then by~\cite[Theorem 4.3]{Wei2020TheoreticalAO},

\begin{align*}
R_i\!\left(f_i^{(t,2)}\right) 
&\le R_i(f^*)
  + C\Bigl(
          2L_1 \Bigl(
              \tfrac{\tau}{\mu - L_2}
              + \sqrt{\tfrac{4\varepsilon_n}{\mu}}
          \Bigr)
        \Bigr)^\alpha
\\
&\le \tfrac{2}{c-1}\, R_i\!\left(h_k^{(t)}\right)
  + \tfrac{2c\rho}{c-1}+C\Bigl(
          2L_1 \Bigl(
              \tfrac{\tau}{\mu - L_2}
              + \sqrt{\tfrac{4\varepsilon_n}{\mu}}
          \Bigr)
        \Bigr)^\alpha
\\[-2pt]
&
\\
&\le \tfrac{2}{c-1}\Bigl(
      \kappa\, R_i\!\left(f_i^{(t,1)}\right)
      + \tfrac{2w_1(\varepsilon_1+\varepsilon_2+c\rho)}{c-1}
    \Bigr)
  + \tfrac{2c\rho}{c-1}
\\[-2pt]
&\qquad {}+ C\Bigl(
          2L_1 \Bigl(
              \tfrac{\tau}{\mu - L_2}
              + \sqrt{\tfrac{4\varepsilon_n}{\mu}}
          \Bigr)
        \Bigr)^\alpha
\qquad\text{(Theorem~\ref{thm:server_improvement})}
\\
&= \tfrac{4w_1 w_2}{(c-1)^2}\, R_i\!\left(f_i^{(t,1)}\right)
  + \tilde{O}\!\left(n^{-1/2}\right)
\\
&\le \tfrac{4w_1 w_2}{(c-1)^2}\, R_i\!\left(f_i^{(t-1,2)}\right)
  + \tilde{O}\!\left(n^{-1/2}\right).
\end{align*}
The last step uses the classification-safe supervised-update condition stated in Section~\ref{sec:framework}: with high probability, the local update from $f_i^{(t-1,2)}$ to $f_i^{(t,1)}$ does not increase the true $0$--$1$ risk on $\cD_i$ beyond the lower-order generalization slack.

Finally, define $\kappa_2=\tfrac{4w_1w_2}{(c-1)^2}$ and let $\zeta \in \tilde{O}(n^{-1/2})$. By the union bound,
$R_i(f_i^{(T,2)})
\;\le\;
\kappa_2^{T-1}\,R_i(f_i^{(1,2)})
\;+\;
\zeta\sum_{m=0}^{T-1}\kappa_2^m
 \le \kappa_2^{T} R_i(f_i^{(1,1)}) + \zeta\sum_{m=0}^{T}\kappa_2^m\;=\;
\kappa_2^{T}\,R_i(f_i^{(1,2)})
\;+\;
\frac{\zeta\bigl(1-\kappa_2^{T}\bigr)}{1-\kappa_2}.
$ Hiding the constants, we complete the proof.
\end{proof}

\section{Additional Experimental Details}
\label{appendix:exp_settings}

\paragraph{Overall Setup:}
Our development environment was based on Python (version 3.10) and PyTorch (version 2.5.0+cu121), with GPU acceleration provided by two NVIDIA GeForce RTX 4090 GPUs.\footnote{The codebase is being prepared for public release.}
All experiments follow a standard cross-device federated learning setting, where a central server coordinates training across many clients with disjoint local datasets.

We measure performance using the classification accuracy of each client on a test set sampled from its own local data distribution, ensuring that evaluation aligns with the personalized nature of the task. For COSMOS, the prediction model used for client $i$ is the \emph{cluster-specific server model} $h_{\pi(i)}$, which represents the distilled knowledge from its assigned cluster. For baseline methods that do not employ a server-side model (e.g., FedMD, FedCT, COMET, and homogeneous PFL methods), accuracy is computed directly using the client's own trained model. This evaluation protocol ensures that all methods are compared according to the model actually deployed to each client.

\paragraph{Resource Requirements:}
Our experiments were conducted on a server running Ubuntu $20.04.6$ LTS, with the development environment based on Python $3.10$ and PyTorch $2.5.0+cu121$. The system was equipped with $2$x NVIDIA GeForce RTX $4090$ GPUs, $94$ GB of RAM, and $2.0$ GB of swap memory.

\paragraph{Time Requirements:} 
When running COSMOS with a VGG architecture using $5$ clusters, training models for one communication round took approximately $20$ minutes. A full experiment (spanning $20$ rounds and $25$ clients) typically required $7$ hours.

For benchmarks with lower complexity (e.g., AlexNets), training models for one communication round took approximately $5$ minutes, and a full experiment (spanning $20$ rounds and $25$ clients) typically required $1.5$ hours.

\paragraph{Data Sources and Details:}
\begin{enumerate}
    \item \textbf{Tiny ImageNet:} The Tiny ImageNet dataset is downloaded from the official Stanford CS231n website. It is provided as a ZIP file containing 200 categories of images, used for classification tasks. The dataset is extracted and reorganized into appropriate folders for training and validation. The dataset is available at: \url{http://cs231n.stanford.edu/tiny-imagenet-200.zip}.
    
    \item \textbf{CIFAR-10 \& CIFAR-100:} These datasets consist of 60,000 32x32 color images in 10 and 100 classes, respectively. They are commonly used for training machine learning models for image classification. The datasets are available for download directly via the torchvision library. More information on these datasets can be found at: \url{https://www.cs.toronto.edu/~kriz/cifar.html}.
    
    \item \textbf{Extended MNIST (EMNIST):} The EMNIST dataset extends the original MNIST dataset to include handwritten letters. We specifically use the \textit{Balanced} version, which includes 131,600 characters across 47 balanced classes. The dataset is available for download via the torchvision library. Additional details can be found on the homepage: \url{https://www.nist.gov/itl/products-and-services/emnist-dataset}.
\end{enumerate}

\paragraph{Hyper Parameters and Optimizers:}

\begin{enumerate}
    \item \textbf{Loss Functions:} During training using pseudo-labels, we use the Kullback-Leibler Divergence (nn.KLDivLoss) to align model outputs with mean pseudo-labels, with reduction='batchmean'. During client fine-tuning (using their local data true labels), we use Cross-Entropy Loss (nn.CrossEntropyLoss).

    \item \textbf{Optimizers and Learning Rate:} We use the Adam optimizer for both training stages. After performing hyperparameter optimization, we use a higher learning rate of $0.001$ when training on client-side data with true labels. This enables the model to adapt quickly and effectively to reliable supervision. In contrast, for pseudo-labeled data, we reduce the learning rate to $0.0001$ to account for the inherent noise and uncertainty in these labels. 


    \item \textbf{Weights Initialization:} For consistency across experiments, we initialize weights using Kaiming He initialization for both convolutional and linear layers. This method, which is well-suited for ReLU activations, helps prevent vanishing or exploding gradients. The biases are initialized to zeros. The seed is updated and applied to ensure reproducibility across runs.

    \item \textbf{COSMOS Clustering Hyperparameter ($B$):}  COSMOS uses a neighborhood threshold hyperparameter $B$  to determine client similarity during the greedy-elimination clustering step. In our synthetic benchmarks, we selected $B$ from a small sweep that often yielded about $K = 5$ clusters, matching the coarse 20\% class-grouping structure used to induce heterogeneous label distributions. We do not view this as prior knowledge of the true cluster count; rather, it reflects a practical tuning choice for the operating point of the greedy clustering rule.
\end{enumerate}

\begin{table}[ht]
\centering
\caption{Final-round performance on benchmark datasets (mean $\pm$ SEM) under
Dirichlet data heterogeneity with $\alpha = 5$.
Rows marked with $^\dagger$ denote that COSMOS's SEM error bar does not overlap
with any baseline (COMET, FedCT, or FedMD). Figure~\ref{fig:diff_benchmarks_05}
presents performance across communication rounds.}
\begin{tabular}{lcccc}
\toprule
Dataset & COMET & FedCT & FedMD & COSMOS \\
\midrule
CIFAR10          & 78.2 $\pm$ 0.25 & 76.9 $\pm$ 0.65 & 74.6 $\pm$ 2.43 & \textbf{81.4 $\pm$ 1.11}$^\dagger$ \\
CIFAR100         & 31.1 $\pm$ 0.65 & 34.5 $\pm$ 0.22 & 24.1 $\pm$ 0.86 & \textbf{44.3 $\pm$ 1.35}$^\dagger$ \\
EMNIST & 87.0 $\pm$ 0.05 & 90.6 $\pm$ 0.30 & 84.1 $\pm$ 0.94 & \textbf{93.9 $\pm$ 0.35}$^\dagger$ \\
TinyImageNet     & 32.2 $\pm$ 1.00 & 35.9 $\pm$ 0.08 & 30.9 $\pm$ 1.81 & \textbf{46.9 $\pm$ 0.29}$^\dagger$ \\
\bottomrule
\end{tabular}

\label{tab:benchmarks-final}
\end{table}

\begin{table}[ht]
\centering
\caption{Final-round performance under different client architecture mixtures (mean $\pm$ SEM).
Rows marked with $^\dagger$ indicate COSMOS's SEM interval does not overlap with \emph{any} baseline. Figure~\ref{fig:diff_clients_nets} presents performance across communication rounds.}
\begin{tabular}{@{}lcccc@{}}
\toprule
Arch. mix & COMET & FedCT & FedMD & COSMOS \\
\midrule
Alex / Mobile             & 24.2 $\pm$ 0.32 & 27.0 $\pm$ 0.48 & 19.3 $\pm$ 1.35 & \textbf{42.1 $\pm$ 0.26}$^\dagger$ \\
Alex / Res                & 32.9 $\pm$ 0.30 & 35.1 $\pm$ 0.37 & 29.8 $\pm$ 1.20 & \textbf{40.3 $\pm$ 0.78}$^\dagger$ \\
Alex / Res / Mobile / Sqz & 32.1 $\pm$ 0.27 & 34.2 $\pm$ 0.35 & 25.9 $\pm$ 1.17 & \textbf{42.5 $\pm$ 1.10}$^\dagger$ \\
Alex / Sqz                & 19.4 $\pm$ 0.72 & 21.1 $\pm$ 0.44 & 13.6 $\pm$ 0.49 & \textbf{37.5 $\pm$ 0.12}$^\dagger$ \\
Mobile / Res / Alex       & 40.8 $\pm$ 0.94 & 42.0 $\pm$ 0.61 & 41.4 $\pm$ 1.47 & \textbf{42.1 $\pm$ 0.53} \\
Mobile / Sqz              & 31.1 $\pm$ 0.65 & 27.2 $\pm$ 0.95 & 24.1 $\pm$ 0.86 & \textbf{44.3 $\pm$ 1.35}$^\dagger$ \\
Res / Mobile              & 24.1 $\pm$ 0.25 & 26.8 $\pm$ 0.31 & 24.6 $\pm$ 0.15 & \textbf{43.9 $\pm$ 1.78}$^\dagger$ \\
Res / Sqz                 & 20.1 $\pm$ 0.69 & 22.4 $\pm$ 0.52 & 19.3 $\pm$ 1.06 & \textbf{40.3 $\pm$ 0.40}$^\dagger$ \\
\bottomrule
\end{tabular}

\label{tab:arch-mixtures-final}
\end{table}

\begin{table}[t]
 \caption{Comparison of Baselines by Communication Type (Bandwidth Measured Per Client in MB).}
    \centering
    \begin{tabular}{lcc}
        \toprule
        \textbf{Dataset} & \textbf{Parameters} & \textbf{Pseudo-labels} \\
        \midrule
        CIFAR-100        & 114.76 & 3.81 \\
        CIFAR-10         & 113.36 & 0.38 \\
        EMNIST-balanced  & 113.93 & 4.04 \\
        Tiny ImageNet    & 116.32 & 15.26 \\
        \bottomrule
    \end{tabular}
   
    \label{table:communication_comparison}
\end{table}

\section{Additional Results}\label{appendix:additional_results}

This appendix provides supplementary empirical results that support and extend the main findings in Section~\ref{S:exp}. The results offer additional evidence regarding robustness, scalability, and design choices in COSMOS.

\paragraph{Effect of Data Heterogeneity:}
\begin{table}[ht]
\centering
\caption{Final-round performance on benchmark datasets (mean $\pm$ SEM) under
Dirichlet data heterogeneity with $\alpha = 1$.
Rows marked with $^\dagger$ denote that COSMOS's SEM error bar does not overlap
with any baseline (COMET, FedCT, or FedMD). Figure~\ref{fig:diff_benchmarks_1}
presents performance across communication rounds.}
\begin{tabular}{lcccc}
\toprule
Dataset & COMET & FedCT & FedMD & COSMOS \\
\midrule
CIFAR10 
& 79.1 $\pm$ 0.98 
& 77.7 $\pm$ 1.01 
& 71.8 $\pm$ 0.07 
& 79.0 $\pm$ 0.11 \\

CIFAR100 
& 42.0 $\pm$ 0.31 
& 39.3 $\pm$ 0.19 
& 29.4 $\pm$ 0.50 
& 43.1 $\pm$ 1.12 \\

EMNIST 
& 86.8 $\pm$ 1.06 
& 91.5 $\pm$ 0.32 
& 83.9 $\pm$ 0.60 
& \textbf{93.5 $\pm$ 0.40}$^\dagger$ \\

TinyImageNet 
& 38.1 $\pm$ 0.20 
& 40.7 $\pm$ 0.45 
& 36.6 $\pm$ 2.38 
& \textbf{46.5 $\pm$ 0.74}$^\dagger$ \\
\bottomrule
\end{tabular}

\label{tab:benchmarks-final_alpha1}
\end{table}

\begin{table}[ht]
\centering
\caption{Final-round performance on benchmark datasets (mean $\pm$ SEM) under
Dirichlet data heterogeneity with $\alpha = 100$.
Rows marked with $^\dagger$ denote that COSMOS's SEM error bar does not overlap
with any baseline (COMET, FedCT, or FedMD). Figure~\ref{fig:diff_benchmarks_100}.
presents performance across communication rounds.}
\begin{tabular}{lcccc}
\toprule
Dataset & COMET & FedCT & FedMD & COSMOS \\
\midrule
CIFAR10
& 79.8 $\pm$ 0.41
& 77.2 $\pm$ 0.15
& 70.0 $\pm$ 1.12
& \textbf{81.4 $\pm$ 0.76}$^\dagger$ \\

CIFAR100
& 39.4 $\pm$ 0.40
& 33.3 $\pm$ 0.52
& 21.6 $\pm$ 0.31
& \textbf{44.8 $\pm$ 1.33}$^\dagger$ \\

EMNIST
& 93.3 $\pm$ 0.05
& 91.2 $\pm$ 0.26
& 85.3 $\pm$ 0.19
& \textbf{93.8 $\pm$ 0.38}$^\dagger$ \\

TinyImageNet
& 37.1 $\pm$ 0.37
& 33.4 $\pm$ 0.25
& 27.3 $\pm$ 2.26
& \textbf{46.2 $\pm$ 0.40}$^\dagger$ \\
\bottomrule
\end{tabular}

\label{tab:benchmarks-final_alpha100}
\end{table}

\begin{figure}[p]
\centering

\includegraphics[width=\textwidth]{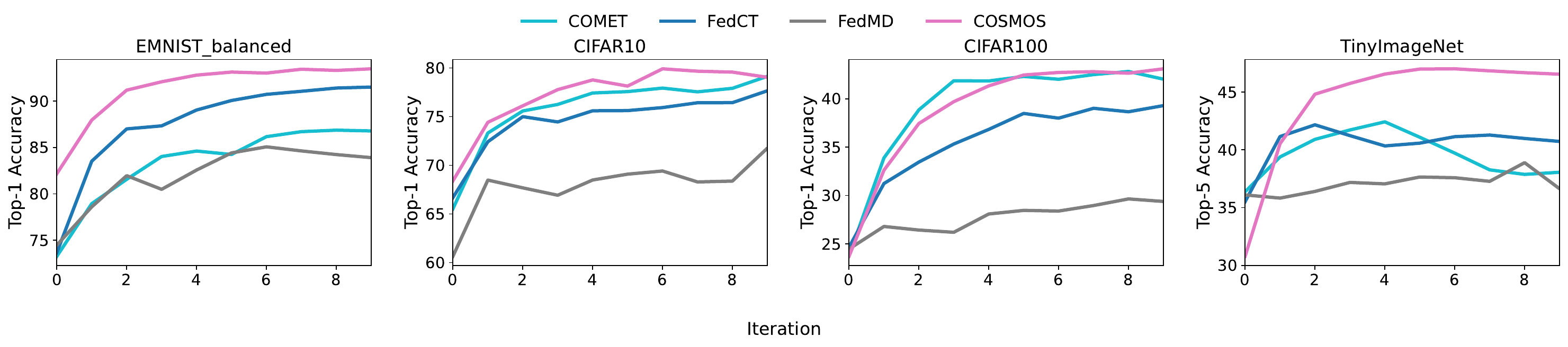}
\caption{Comparison of COSMOS and heterogeneous-model baselines across four benchmarks under Dirichlet non-IID sampling ($\alpha = 1$). Client models consist of MobileNet and SqueezeNet architectures. Curves show mean client accuracy over rounds (Top-1 for  CIFAR-10/100 and EMNIST; Top-5 for Tiny ImageNet). A statistical comparison of final-round performance is provided in Table~\ref{tab:benchmarks-final_alpha1}.}
\label{fig:diff_benchmarks_1}
\end{figure}

\begin{figure}[p]
\centering

\includegraphics[width=\textwidth]{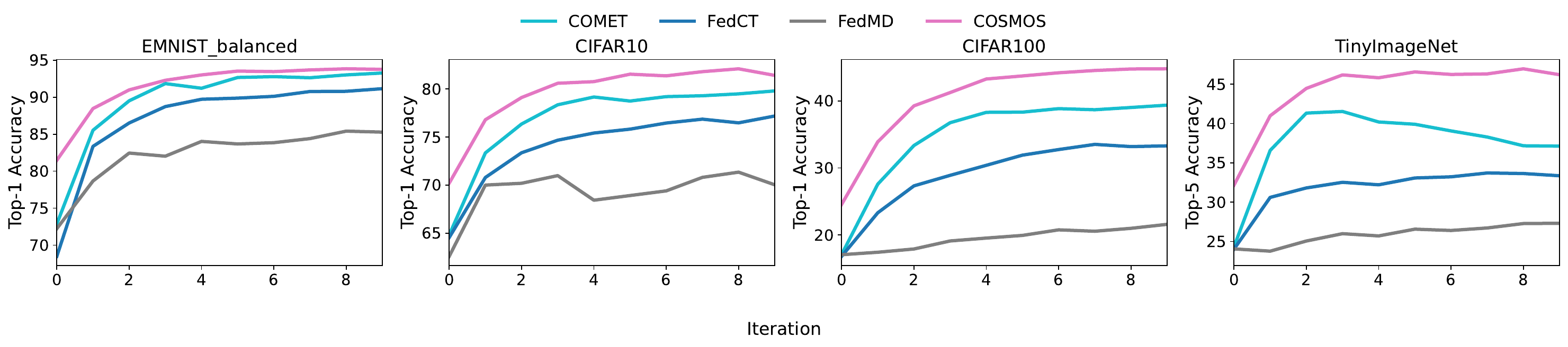}
\caption{Comparison of COSMOS and heterogeneous-model baselines across four benchmarks under Dirichlet non-IID sampling ($\alpha = 100$). Client models consist of MobileNet and SqueezeNet architectures. Curves show mean client accuracy over rounds (Top-1 for  CIFAR-10/100 and EMNIST; Top-5 for Tiny ImageNet). A statistical comparison of final-round performance is provided in Table~\ref{tab:benchmarks-final_alpha100}.}
\label{fig:diff_benchmarks_100}
\end{figure}

In Figures~\ref{fig:diff_benchmarks_1} and~\ref{fig:diff_benchmarks_100}, as well as Tables~\ref{tab:benchmarks-final_alpha1} and~\ref{tab:benchmarks-final_alpha100}, we examine the impact of data heterogeneity by varying the Dirichlet parameter $\alpha$. Across all datasets, COSMOS consistently outperforms model-agnostic baselines for all evaluated values of $\alpha$. While the absolute performance of all methods degrades as heterogeneity increases, the relative advantage of COSMOS becomes more pronounced in highly non-IID regimes.

\paragraph{Comparison in Setting with Homogeneous Clients:}
\begin{table}[ht]
\caption{Final-round performance under homogeneous client architectures (mean $\pm$ SEM). Best result per architecture in bold.}
\label{tab:homogeneous-architectures}
\centering

\begin{tabular}{lcccc}
\toprule
Method & AlexNet & ResNet & MobileNet & SqueezeNet \\
\midrule
COMET     & 30.1 $\pm$ 0.46 & 35.1 $\pm$ 0.25 & 35.5 $\pm$ 0.11 & 24.9 $\pm$ 0.18 \\
Ditto     & 26.8 $\pm$ 1.48 & 34.2 $\pm$ 0.40 & 34.5 $\pm$ 0.14 & 25.7 $\pm$ 0.22 \\
FedBABU   & 7.2 $\pm$ 0.14  & \textbf{47.9 $\pm$ 0.17} & \textbf{49.1 $\pm$ 0.21} & 18.5 $\pm$ 0.67 \\
FedMD     & 19.5 $\pm$ 0.46 & 35.8 $\pm$ 0.27 & 29.3 $\pm$ 0.08 & 15.8 $\pm$ 0.63 \\
FedSelect & 25.6 $\pm$ 0.15 & 27.9 $\pm$ 0.24 & 19.7 $\pm$ 0.30 & 16.2 $\pm$ 0.26 \\
COSMOS    & 25.7 $\pm$ 11.57 & 44.8 $\pm$ 1.41 & 46.6 $\pm$ 1.53 & \textbf{35.5 $\pm$ 0.30} \\
pFedCK    & 28.1 $\pm$ 0.00 & 40.3 $\pm$ 0.26 & 33.9 $\pm$ 0.15 & 30.5 $\pm$ 0.23 \\
pFedHN    & \textbf{33.4 $\pm$ 0.31} & 27.9 $\pm$ 0.26 & 21.9 $\pm$ 0.89 & 23.1 $\pm$ 0.38 \\
\bottomrule
\end{tabular}
\end{table}

\begin{figure}[h]
\centering

\includegraphics[width=\textwidth]{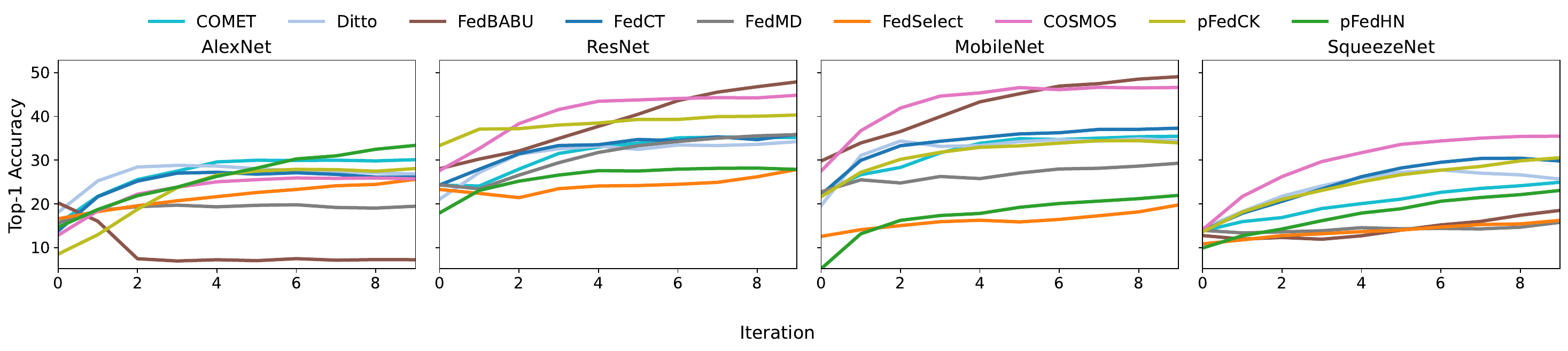}
\caption{Top-1 accuracy over communication rounds for homogeneous client architectures (AlexNet, ResNet, MobileNet, and SqueezeNet) on CIFAR-100 with Dirichlet non-IID partitioning ($\alpha = 5$). Each curve shows the mean accuracy across seeds for a variety of algorithms that do not require clients to be model-agnostic: COMET, Ditto, FedBABU, FedMD, FedSelect, COSMOS, pFedCK, and pFedHN. A statistical summary of the final communication round appears in Table~\ref{tab:homogeneous-architectures}.}
\label{fig:homog_arch}
\end{figure}
 
In Table~\ref{tab:homogeneous-architectures} and Figure~\ref{fig:homog_arch}, we relax the model-agnostic constraint and compare COSMOS to additional baselines in a homogeneous setting where all clients share the same architecture. COSMOS remains competitive with state-of-the-art personalized FL methods, while no single baseline dominates across all datasets and architectures. This shows that COSMOS's model-agnostic design does not come at the cost of performance when homogeneity is available.

\paragraph{Additional Comparisons across Various Settings:}

\begin{figure}[p]
\centering
\includegraphics[width=\columnwidth]{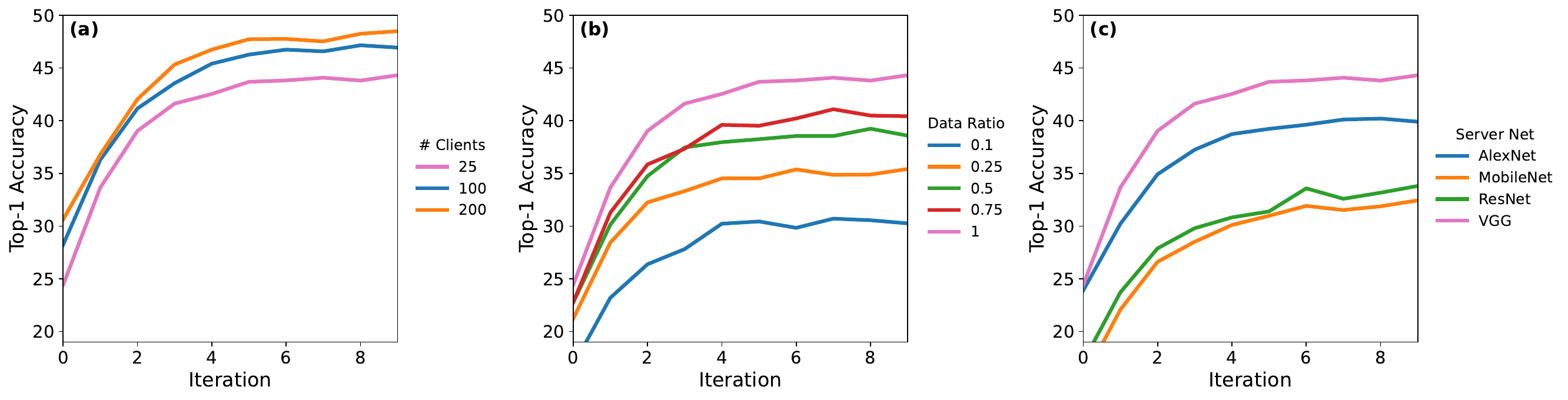}
\caption{Top-1 accuracy over communication rounds for COSMOS on CIFAR-100 with Dirichlet non-IID sampling ($\alpha = 5$): 
(a) varying numbers of participating clients, 
(b) different ratios (fractions) of unlabeled global data, and 
(c) different server-side architectures. 
Client models are MobileNet and SqueezeNet.}
\label{fig:client-scale}
\end{figure}

\begin{table}[ht]
\centering
\caption{Final-round performance of COSMOS under: (a) varying numbers of participating clients, (b) different ratios (fractions) of unlabeled global data, and (c) different server-side architectures.}
\begin{tabular}{lc}
\toprule
Setting & COSMOS (final-round, mean $\pm$ SEM) \\
\midrule
\multicolumn{2}{l}{\textbf{(a) \# Clients}} \\
25  & 44.3 $\pm$ 1.35 \\
100 & 46.9 $\pm$ 0.44 \\
200 & 48.5 $\pm$ 0.41 \\
\midrule
\multicolumn{2}{l}{\textbf{(b) Unlabeled global data ratio}} \\
0.10 & 30.3 $\pm$ 0.55 \\
0.25 & 35.4 $\pm$ 0.95 \\
0.50 & 38.6 $\pm$ 0.29 \\
0.75 & 40.4 $\pm$ 0.68 \\
1.00 & 44.3 $\pm$ 1.35 \\
\midrule
\multicolumn{2}{l}{\textbf{(c) Server architecture}} \\
AlexNet   & 39.9 $\pm$ 0.27 \\
MobileNet & 32.4 $\pm$ 1.16 \\
ResNet    & 33.8 $\pm$ 0.99 \\
VGG       & 44.3 $\pm$ 1.35 \\
\bottomrule
\end{tabular}

\label{tab:clientscale-globalratio-servernet-final}
\end{table}

We next evaluate the performance of COSMOS under different settings in Figure~\ref{fig:client-scale} and Table~\ref{tab:clientscale-globalratio-servernet-final} by varying (a) the number of participating clients, (b) the size of the unlabeled public dataset $U$, and (c) the server-side model architecture. As the number of clients increases, COSMOS continues to improve, benefiting from increased pseudo-label diversity at the server. Similarly, increasing $|U|$ results in consistent performance gains. Figure~\ref{fig:client-scale}(c) shows that COSMOS further benefits from stronger server-side models.

\paragraph{Comparison with Different Numbers of Clusters:}

\begin{figure}[h]
\centering
\includegraphics[width=0.6\columnwidth]{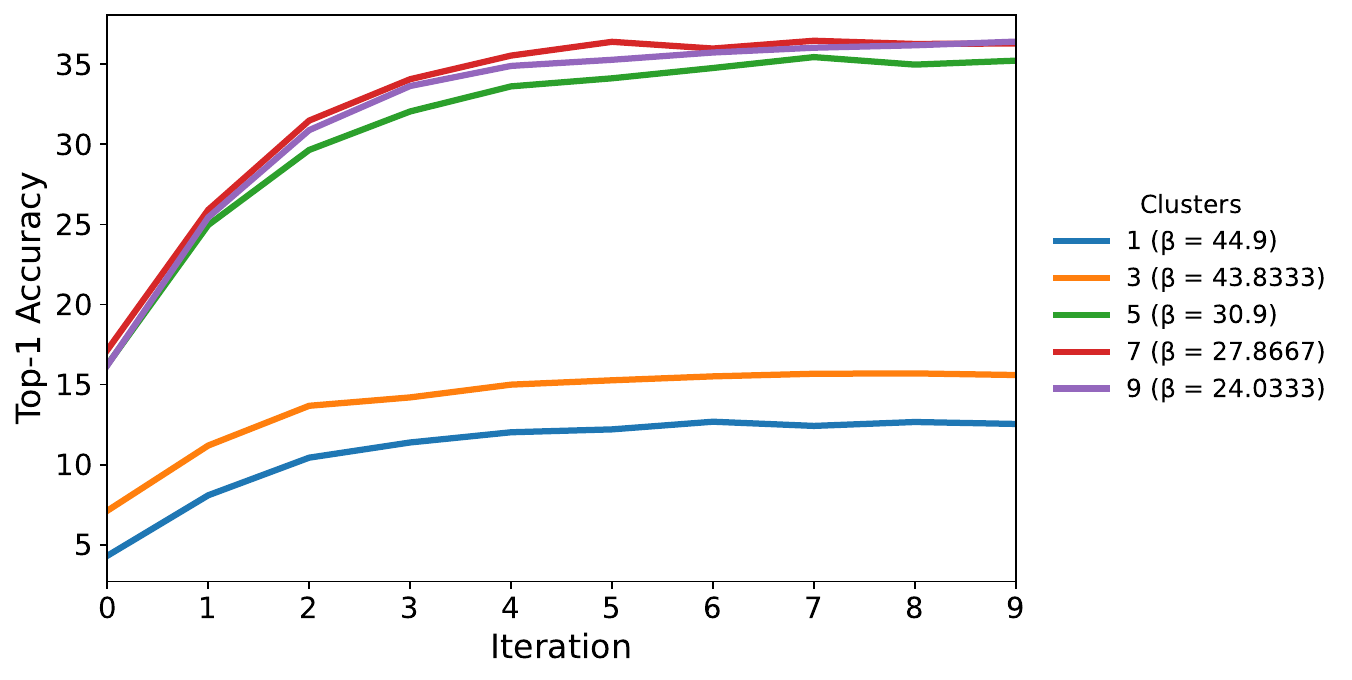}
\caption{Top-1 accuracy over communication rounds for COSMOS showing the performance on CIFAR-100 with Dirichlet non-IID sampling ($\alpha = 5$) under varying number of clusters (i.e., values of $\beta$ shown in the legend), Client models are AlexNet.} 

\label{fig:COSMOS_cluster}
\end{figure}

In Figure~\ref{fig:COSMOS_cluster}, we study the effect of tuning $\beta$, where different values correspond to different numbers of clusters. We observe that performance saturates once the number of clusters reaches approximately five, with no further improvement beyond this point. In contrast, larger values of $\beta$ (resulting in fewer clusters) lead to degraded performance, likely because grouping too many clients together reduces data similarity within each cluster.

\begin{figure}[t]
\centering
\includegraphics[width=\columnwidth]{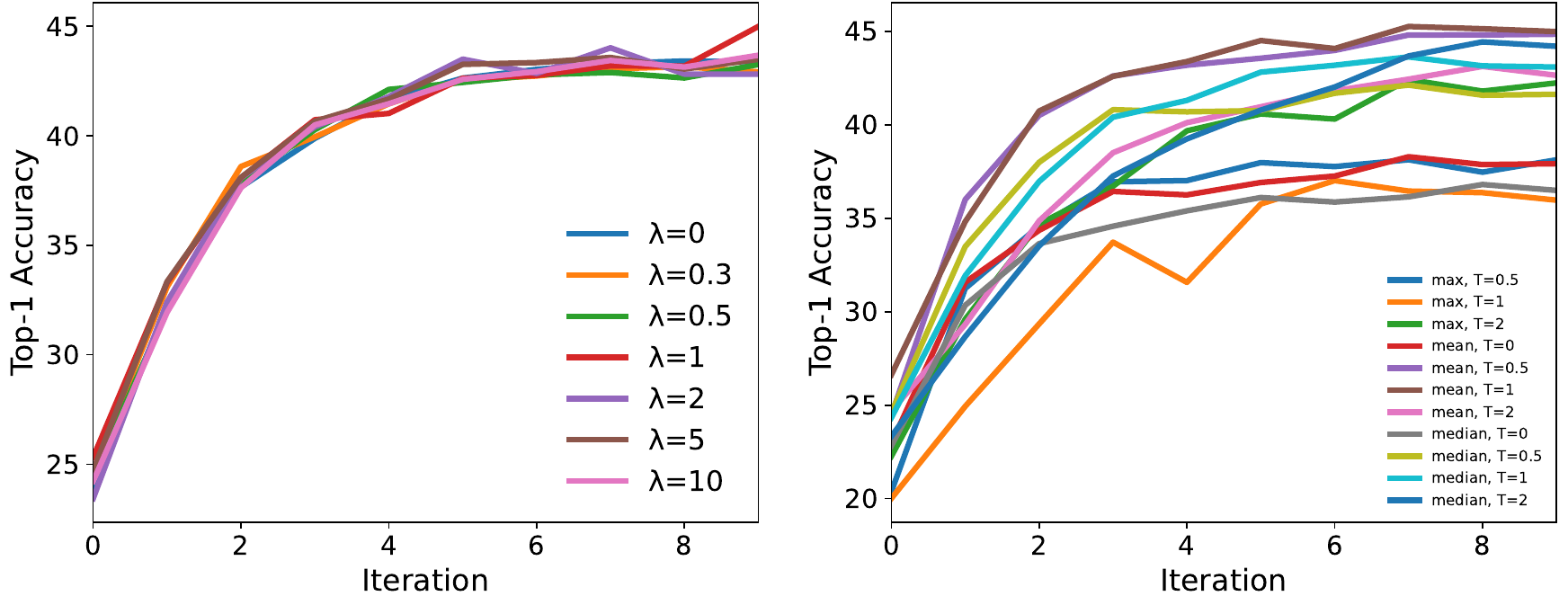}
\caption{Top-1 accuracy over communication rounds for COSMOS. Subfigure (a) shows performance under different values of $\lambda$. Subfigure (b) shows performance on CIFAR-100 with Dirichlet non-IID sampling ($\alpha = 5$) under varying distillation temperatures and different statistical aggregation rules (mean, median, max). Client models are MobileNet and SqueezeNet.} 

\label{fig:COSMOS_params}
\end{figure}

\paragraph{Comparison with Different Hyperparameter Values:}
Lastly, in Figure~\ref{fig:COSMOS_params}, we examine the performance of COSMOS under different parameter settings. In subfigure (a), we vary the value of $\lambda$, while in subfigure (b) we evaluate different pseudo-label aggregation functions within each cluster, as well as the effect of pseudo-label temperature (with temperature $0$ corresponding to hard labels). COSMOS is robust to moderate variations in $\lambda$, with performance remaining stable across a wide range of values and $\lambda = 5$ yielding the best overall results in our experiments. We also observe that $T = 1$ performs best across datasets, while larger temperatures tend to oversmooth pseudo-labels and slightly degrade performance. This indicates that preserving confident predictions during pseudo-label exchange is beneficial. Additionally, mean aggregation consistently achieves the strongest performance among the evaluated aggregation strategies.


For completeness, we provide statistical comparisons of final-round performance using mean accuracy $\pm$ SEM across three random seeds. In most evaluated settings, COSMOS exhibits non-overlapping error bars with model-agnostic baselines, providing strong evidence of consistent performance improvements.

\end{document}